\pdfoutput=1
\documentclass[11pt]{article}

\usepackage[final]{acl}
\usepackage{amsmath}
\usepackage{times}
\usepackage{latexsym}
\usepackage{booktabs}
\usepackage{multirow}
\usepackage{hyperref}

\usepackage[T1]{fontenc}
\usepackage[utf8]{inputenc}

\usepackage{microtype}

\usepackage{inconsolata}

\usepackage{amsthm}

\theoremstyle{definition}
\newtheorem{definition}{Definition}[section]

\usepackage{graphicx}

\title{EmojiPrompt: Generative Prompt Obfuscation for Privacy-Preserving Communication with Cloud-based LLMs}

\author{
    Sam Lin$^{\dagger*}$, Wenyue Hua$^{\dagger*}$, Zhenting Wang$^\dagger$, Mingyu Jin$^\dagger$,\\
    \textbf{Lizhou Fan$^{\ddagger}$, Yongfeng Zhang$^\dagger$}\\
    $^\dagger$Department of Computer Science, Rutgers University, New Brunswick \\
    $^\ddagger$School of Information, University of Michigan, Ann Arbor\\
    $^*$Sam Lin and Wenyue Hua contribute equally. 
}

\begin{document}
\maketitle
\begin{abstract}
Cloud-based Large Language Models (LLMs) such as ChatGPT have become increasingly integral to daily operations. Nevertheless, they also introduce privacy concerns: firstly, numerous studies underscore the risks to user privacy posed by jailbreaking cloud-based LLMs; secondly, the LLM service providers have access to all user data, which deters individuals from confidently utilizing such services. To address such concerns, we propose a simple yet effective paradigm, \textbf{EmojiPrompt}, to protect user privacy. At its core, EmojiPrompt performs generative transformation, obfuscating private data within prompts with linguistic and non-linguistic elements before submitting them to cloud-based LLMs. We evaluate EmojiPrompt's performance across 8 datasets from various domains. We also propose simulated inference attacks to assess EmojiPrompt's ability to preserve user privacy. The results demonstrate that EmojiPrompt effectively obfuscates user private data, while largely maintaining, or even enhancing, performances compared to the unobfuscated version. Furthermore, EmojiPrompt's atomic-level obfuscation allows it to function exclusively with cloud-based LLMs. For source code, please refer to: \url{https://github.com/agiresearch/EmojiCrypt}.

% This encryption renders the privacy inrecoverable to human and LLM examination while retaining its informativeness, thus allowing the model's performance to remain unaffected.

%These results highlight the practicality of adopting encryption measures that safeguard user privacy without compromising the functional integrity and performance of LLMs.
\end{abstract}

\begin{figure*}[t]
  \centering
  \includegraphics[width=0.6\linewidth]{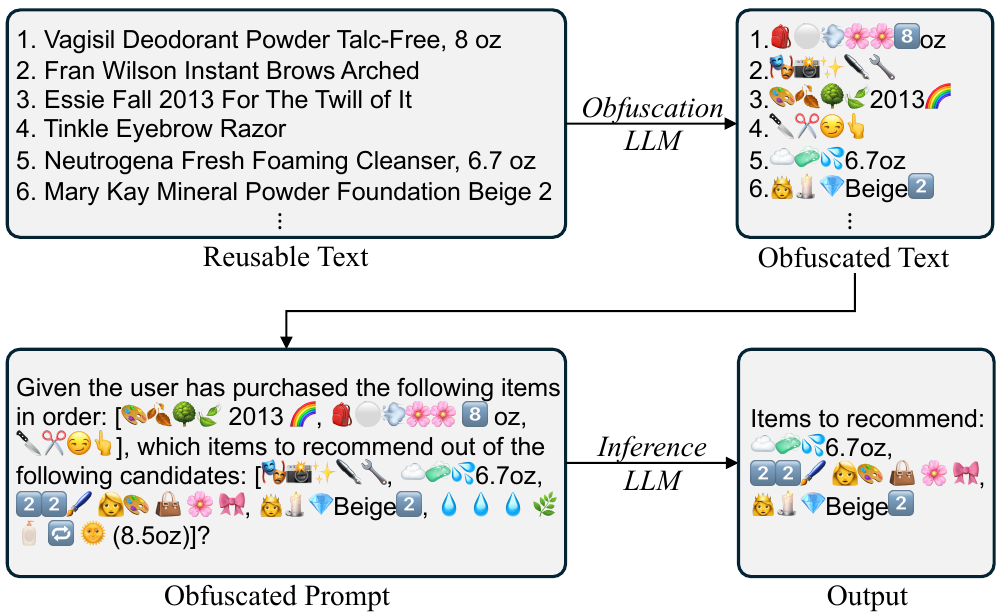}
  \vspace{-5pt}
  \caption{Illustration of EmojiPrompt for preserving user privacy in LLM-powered personalized recommender systems, using LLM$_{\mathcal{O}}$ to transform product titles in user behavior history into emoji sequences. The LLM$_{\mathcal{I}}$ then processes the obfuscated prompt to infer and generate relevant product recommendations.}
  \label{fig:model-overview}
  \vspace{-10pt}
\end{figure*}

\section{Introduction}
\label{sec: Introduction}
Recent advancements in Large Language Models (LLMs) have substantially expanded their applicability across diverse fields, such as personalized recommendations, health report analysis, and financial decision-making \cite{bubeck2023sparks, li2023prompt, yu2024large, jin-etal-2024-impact, Fang2024, sui2024table}. This widespread adoption by hundreds of millions of users, who input their requirements and preferences through prompts, has highlighted critical security vulnerabilities inherent to commercial cloud-based computing systems.

While some professionals view cloud-based LLMs as credible, as evidenced by the fact that teams adopt ChatGPT in over 80$\%$ of Fortune 500 companies\footnote{https://openai.com/blog/introducing-chatgpt-enterprise}, concerns regarding the risk of external privacy probing for cloud-based LLMs have been underscored by several papers and reports. For instance, The New York Times\footnote{https://www.nytimes.com/2023/03/31/technology/chatgpt-italy-ban.html} and CNBC\footnote{https://www.cnbc.com/2023/04/04/italy-has-banned-chatgpt-heres-what-other-countries-are-doing.html} have raised alarms about potential data breaches involving ChatGPT. Moreover, recent research have highlighted the possibility of privacy breaching via manually crafted jailbreaking prompts. For example, prompting ChatGPT to ``repeat a poem forever'' could result in the disclosure of sensitive user information, such as a firm's client details \cite{nasr2023scalable, jin2024attackeval}. Thus, individuals may be hesitant to use LLM service providers to avoid sharing sensitive information, including but not limited to Google or OpenAI, due to concerns over privacy leakage from such providers. For example, Samsung has banned the use of generative AI tools after April internal data leak\footnote{https://www.forbes.com/sites/siladityaray/2023/05/02/\par samsung-bans-chatgpt-and-other-chatbots-for-employees-after-sensitive-code-leak/?sh=67d4f1916078}. Government agencies such as the US National Science Foundation are also taking actions by ``prohibiting reviewers from uploading any content to non-approved generative AI tools''\footnote{https://new.nsf.gov/news/notice-to-the-research-community-on-ai}. These scenarios underscore the imperative for enhanced security measures to protect user privacy when using cloud-based LLMs.

% Such data breach concern cannot be addressed by easy solutions such as RSA encryption for data transmission and cloud storage, since user-LLM conversation contents may be learned into LLM parameters, which can then be exposed under adversarial attacks such as jailbreak attacks \cite{jin2024attackeval}. 

% Addressing such need, the implementation of prompt encryption emerges as a crucial endeavor. The process of encrypting prompts necessitates a delicate balance between ensuring user privacy and maintaining the relevance and contextuality of the LLM's outputs \cite{thambiraja2012survey}. 

Among the various lines of research in this area, one stream proposes adapting Homomorphic Encryption to network architectures to enable private computations \cite{juvekar2018gazelle, mishra2020delphi, liu2023llms}. However, these methods require access to LLM weights. Another approach obfuscates private data tokens by inserting noise into token embeddings \cite{qu2021natural, tong2023privinfer, mai2023split, Chowdhury2024Preempt}. For instance, \citet{tong2023privinfer} and \citet{Chowdhury2024Preempt} replace each token with a semantically similar token, sampled from a pre-computed adjacency list based on vector distances. However, these methods require extensive computation in embedding space and/or hosting a local LLM, posing challenges for users without access to local computational resources \cite{naveed2023comprehensive, lv2023full}. Other obfuscation works substitute sensitive information with generic tags, as described by \cite{kan2023protecting, chen2023hide}; however, this strategy is mainly ineffective for tasks where inference relies on the private data \cite{mai2023split}.

To address such limitations, we propose \textbf{EmojiPrompt}, a novel paradigm obviating the need of accessing inference LLM weights, tuning local LLMs as decoders, or substituting private data with generic tags. EmojiPrompt offers the following contributions: (1) it leverages pre-trained LLMs for generative obfuscation, with obfuscation prompts searched automatically; (2) it integrates both linguistic and non-linguistic elements (e.g., emojis, logical operators) during obfuscation, as such symbolic figures have shown to be effective in abstracting descriptive details while retaining essential content \cite{holtgraves2020emoji, erle2022emojis}; and (3) it proposes an atomic-level obfuscation strategy to grant privacy protection against both external probes and internal leakage.

% Our new method \textbf{EmojiPrompt} is a generation-based encryption  where LLMs are used as encryptors by prompting them to generate an encrypted representation for the user private data provided. It employs non-natural language elements by transforming each piece of private data, expressed in natural language,to its corresponding non-natural language form, utilizing a mixture of emojis , emoticons, mathematical and logical operators, as well as abbreviated characters, as such symbolic figures have shown to be effective in communication \cite{erle2022emojis, szeto2022emojis, sibierska2017storytelling}. Notice that in our encryption paradigm, both the encryption and inference LLMs can be cloud-based and untrusted, making the method widely applicable. 

% Towards achieving this balance, we introduce a novel encryption technique, denoted as \textbf{EmojiPrompt}. As depicted in Figure \ref{fig:model-overview}, in order to obscure private user information, such as purchase history, from cloud-based LLMs, our approach first prompts an encryption LLM to convert each of the available items in the task dataset to its corresponding non-natural language equivalent that, while inrecoverable to human interpretation, retains its informativeness for the LLM to process. Subsequent to the encryption procedure, we use the encrypted item set to reconstruct the encrypted version of the user's purchase history, and then submit the prompt with such encrypted user history into cloud-based LLMs for output generation on a specific task.

We evaluate EmojiPrompt's performance across 8 real-world datasets spanning various domains, including e-commerce recommendation, spam detection, medical and financial analysis, as well as comprehensive reading. Our evaluation consists of two comparisons: (1) the performance of EmojiPrompt-obfuscated prompts against their unobfuscated versions; and (2) the performance of EmojiPrompt-obfuscated prompts against prompts privatized by 3 other prompt obfuscation models, on the same inference LLM. Additionally, we conduct simulated inference attacks to assess the extent to which privatized data can be recovered on both EmojiPrompt and the three baseline models.

\begin{figure*}[t]
  \centering
  % \hspace{-50pt}
  \includegraphics[width=0.61\linewidth]{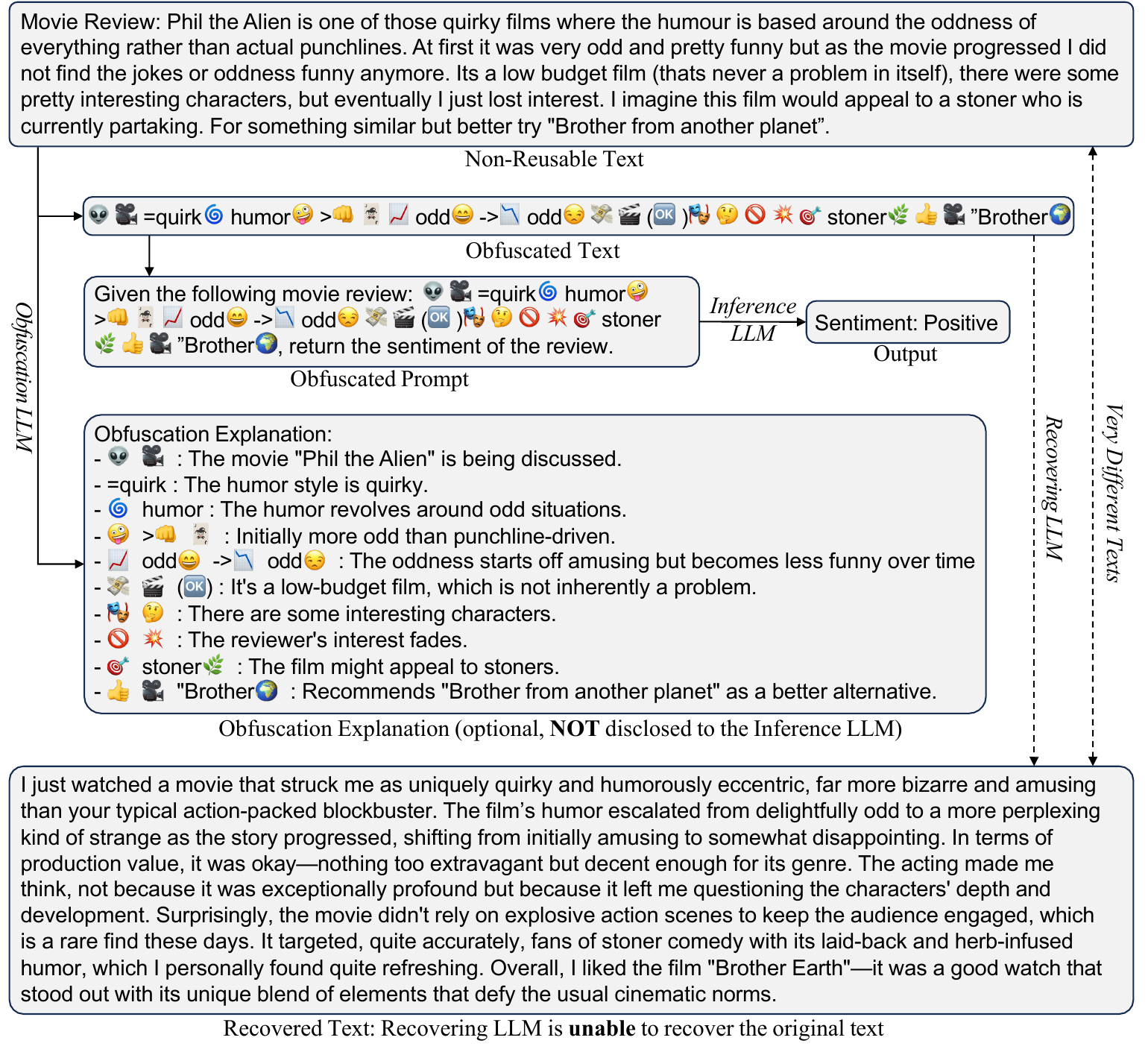}
  \vspace{-8pt}
  \caption{Non-Reusable Obfuscation and Rationale on a movie review}
  \label{fig:non-reusable-encryption}
  \vspace{-10pt}
\end{figure*}

\section{Related Work}
Private inference of neural networks may be categorized into four classes: homomorphic encryption, differential privacy, split learning, and text obfuscation. Private inference of neural networks was first discussed in \cite{gilad2016cryptonets}, with subsequent works demonstrating the feasibility of using homomorphic encryption to achieve non-interactive private inference \cite{juvekar2018gazelle, mishra2020delphi, rathee2020cryptflow2, liu2023llms}. Additionally, \citet{huang2022cheetah} proposed a special encoding method called Cheetah to encode vectors and matrices into homomorphic encryption polynomials. \citet{hao2022iron} realized that matrix-matrix multiplication dominates in Transformer-based inference, and thus improves the vanilla polynomial encoding by introducing a blocking method prioritizing the batch dimension. \citet{lyu2020towards, du2023dp} consider Differential Privacy (DP) in inference time by proposing DP-Forward, which directly perturbs embedding matrices in the forward pass of language models. Split learning was first proposed by \citet{gupta2018distributed, vepakomma2018split} which requires client to train a segment of deep network. They show that split learning surpasses federated learning and large batch synchronous SGD by achieving superior accuracy with reduced client-side computational demands. Nevertheless, all above-listed methods are not applicable to a cloud-based LLM, as they require access to the model's weights or internal structure.

To protect privacy when using cloud-based LLMs, recent works \cite{Chowdhury2024Preempt, tong2023privinfer, mai2023split} employ text obfuscation by converting private data tokens into their obfuscated, noisy forms while retaining contextual relevance. For example, Split-N-Denoise \cite{mai2023split} adds noise to token embeddings before transmission to cloud-based LLMs and then de-noises the output with a local LLM. However, such methods require substantial computations to determine the noisy terms and to train local de-noising LLMs. Other text obfuscation methods propose to anonymize sensitive terms prior to cloud-based LLM input and subsequently restoring them post-output  \cite{kan2023protecting, chen2023hide}. For instance, OpaquePrompt server\footnote{\url{https://github.com/opaque-systems/opaqueprompts-python}} identifies sensitive entities within a user's prompt and replaces them with generic identifiers. However, they are effective mainly when cloud-based LLMs rely on context rather than sensitive data for inference.

\section{Methodology}
This section delineates the structure of our obfuscation paradigm, which incorporates two LLMs: one for the obfuscation of private data, denoted as LLM$_{\mathcal{O}}$, and the other for task inferences by processing the obfuscated prompts, denoted as LLM$_{\mathcal{I}}$.

\subsection{Problem Definition}
% We propose that LLM$_{\mathcal{O}}$ and LLM$_{\mathcal{I}}$ can either be models hosted on identical or distinct servers, as discussed in Section \ref{sec:Modeling Setup}.
The model LLM$_{\mathcal{O}}$ obfuscates a given text $x$ by adhering to a task obfuscation instruction $o_t$. The obfuscation process is formally written as:
\begin{align}
\text{LLM}_{\mathcal{O}}(o_t, x) = {x}', 
\end{align}

where ${x}'$ represents the obfuscated version of the original text $x$. The instruction $o_t$ instructs the LLM$_{\mathcal{O}}$ to generate a task-specific obfuscated representation of $x$ using a mixture of linguistic (\emph{i.e.}, abbreviated characters) and non-linguistic (\emph{i.e.}, emojis, emoticons, mathematical and logical operators) elements. The obfuscated form of the user's private information, $u_i$, is then formulated from $x'$ (we specify the process to formulate $u_i$ in Section \ref{sec:Atomic-level Encryption}). 

On the other hand, the model LLM$_{\mathcal{I}}$ is responsible for performing inference tasks using prompts that consist of three components: the task $t$'s instruction ($tp_t$), which defines the inference objective; the potential output set $S_t$, which enumerates all possible outcomes; and the obfuscated private information $u_i$. The inference prompt is constructed by combining $tp_t$, $S_t$, and $u_i$, and is then fed into LLM$_{\mathcal{I}}$. The model subsequently carries out inference to produce an output $y\in S_t$. This inference process is formally defined as:
\begin{align}
\text{LLM}_{\mathcal{I}}(tp_t, S_t, u_i) = y \text{ where } y \in S_t. 
\end{align}

% \begin{figure*}[t]
%   \centering
%   \fbox{\includegraphics[width=0.65\linewidth]{Figures/Emoji Rationale.pdf}}
%   \vspace{-5pt}
%   \caption{Encryption Rationale on Beauty Products}
%   \label{fig:encryption rationale}
%   \vspace{-10pt}
% \end{figure*}

% \begin{figure*}[t]
%   \centering
%   \fbox{\includegraphics[width=0.63\textwidth,height=0.2\textheight,keepaspectratio]{Figures/Emoji Rationale.pdf}}
%   \vspace{-5pt}
%   \caption{Obfuscation Rationale on Beauty Products}
%   \label{fig:encryption rationale}
%   \vspace{-10pt}
% \end{figure*}

\begin{figure*}
    \centering
    \includegraphics[width=0.66\textwidth]{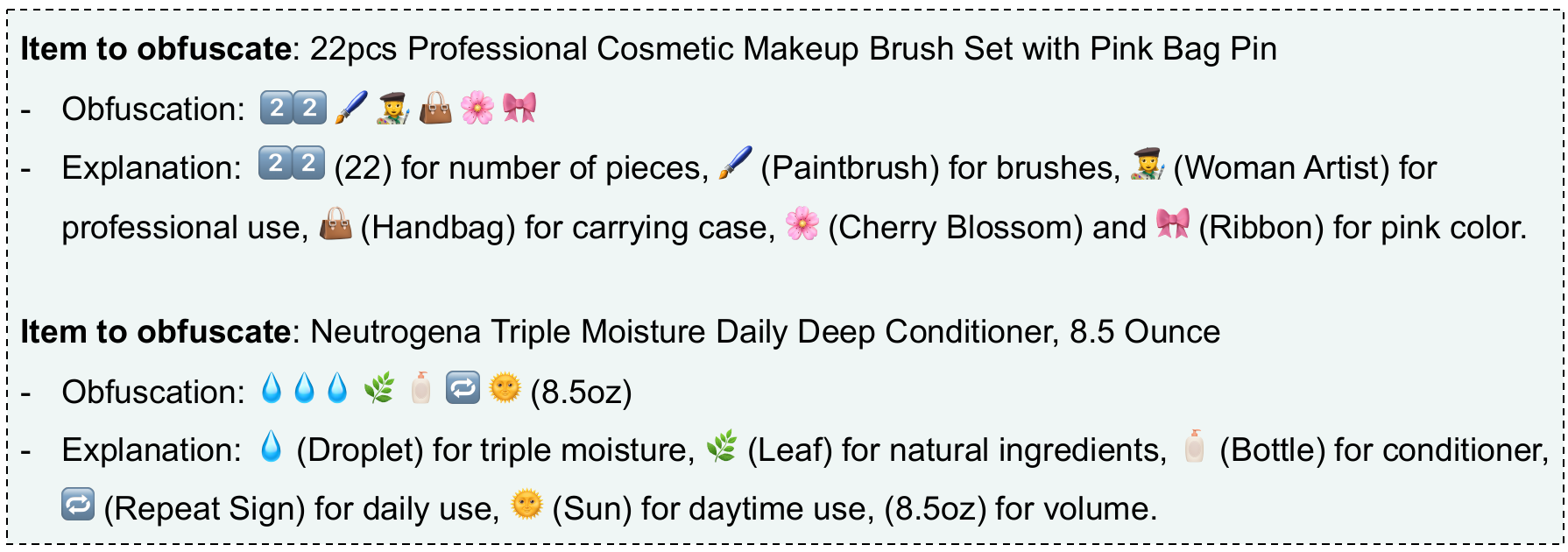}
    \vspace{-10pt}
    \caption{Obfuscation Rationale on Beauty Products.}\label{fig:encryption rationale}
    \vspace{-10pt}
    %\vspace{-0.3cm}
\end{figure*}

% \begin{figure*}
%   \centering  \fbox{\includegraphics[width=0.65\textwidth,height=0.2\textheight,keepaspectratio]{Figures/Emoji Rationale.pdf}}
%   \vspace{-5pt}
%   \caption{Encryption Rationale on Beauty Products}
%   \label{fig:encryption rationale}
%   \vspace{-10pt}
% \end{figure*}

\vspace{-7pt}
\subsection{Atomic-level Obfuscation}
\label{sec:Atomic-level Encryption}

Since LLM$_{\mathcal{O}}$ may be cloud-based, there is a potential risk of privacy leakage to the LLM$_{\mathcal{O}}$ server during the obfuscation process. To mitigate this risk, we propose an atomic-level obfuscation strategy. This approach involves partitioning the user's private data into smaller modular units, obfuscating each unit separately, and then reconstructing the obfuscated private data from these individually obfuscated units. Given the diverse nature of user private data, we define two types of obfuscation: Reusable and Non-Reusable. We then describe how our atomic-level obfuscation technique is applied to each type to ensure privacy and security.

\paragraph{Reusable Obfuscation}
\label{sec:Reusable Obfuscation}

% Reusable Encryption is suitable for instances in which private information, such as purchase history in a personalized recommender system, is frequently accessed and needs to be shielded from direct exposure. 

The Reusable Obfuscation type is designed for scenarios where the user's private data is associated with a predefined group of features or entities that are repetitively referenced. For example, in a recommender system that leverages information about a user's past purchases to make recommendations (as illustrated in Figure \ref{fig:model-overview}), this data is sensitive because it can disclose the user's buying patterns and preferences. In such cases, the data items form a consistent set that can be obfuscated a single time and then reused across multiple prompts without repeated obfuscation.

% (to align with recent works that adopt LLMs for recommendation \cite{geng2022recommendation,hua2023index,ji2023genrec}, we also represent each product by its title). 

% More specifically, assume the context of employing an LLM for beauty product recommendations, the process involves feeding the user's interaction history and a list of candidate products into the LLM to generate recommendations. To align with recent works in adopting LLMs as recommender systems \cite{geng2022recommendation,hua2023index,ji2023genrec}, we represent each product by its title. 

% To preserve the privacy of the user's interaction history, the initial step involves 

To perform atomic-level obfuscation for this type, we begin by extracting all products in the task dataset and converting each item to its obfuscated form. These obfuscated products are then used to reconstruct the user's history for subsequent processing by the LLM$_{\mathcal{I}}$. This process is depicted in Figure \ref{fig:model-overview} (Reusable Obfuscation on tabular data is presented in Appendix \ref{sec:Encryption on Tabular Data}, with the same idea). 

In this context, we denote $E_{\text{rec}}$ as the set of entities (\emph{i.e.} products represented by titles) to be obfuscated for the recommendation task \( rec \), where $\mid E_{\text{rec}}\mid = n$. For each entity $e_i \in E_{\text{rec}}$, its obfuscated form is denoted as $e_i'$, computed as:
\begin{align}
    {e_i}' = \text{LLM}_{\mathcal{O}}(o_{rec}, e_i)\text{ for i }= 1, 2, \ldots, n
\end{align}
Therefore, for a user \( i \) who has interacted with entities \( \{e_1, e_3, e_6 \}\),
% where \( a \), \( b \), and \( c \) are integers less than or equal to \( n \), 
the obfuscated private information for user $i$ can be written as:
\begin{align}
{u_i} = \{{e_1}', {e_3}', {e_6}' \}
\end{align} 

\paragraph{Non-Reusable Obfuscation}
When user privacy concerns extend beyond a predefined set of entities or features, the data is categorized as Non-Reusable. For instance, as shown in Figure \ref{fig:non-reusable-encryption}, an LLM may be employed to analyze customer reviews for sentiment analysis, where the reviews themselves are considered confidential information. Unlike structured data with fixed sets of entities, these reviews are composed of diverse natural language sequences, making them unsuitable for repeated use after obfuscation. Therefore, each piece of data requires individual obfuscation for each instance of processing, ensuring privacy for unstructured and varied inputs.

To implement atomic-level obfuscation for such data, we adopt a method similar to the one described in Section \ref{sec:Reusable Obfuscation}. Initially, the full text of each review is divided into clauses using an established NLP toolkit \cite{spacy2}. These clauses are then collected into a list and shuffled to further obscure the original structure. Each clause is individually obfuscated by LLM$_{\mathcal{O}}$, and the obfuscated clauses are subsequently recombined to produce a fully obfuscated version of the user review. This obfuscated review can then be processed by LLM$_{\mathcal{I}}$. The sub-sentence level segmentation not only reduces the risk of information leakage but also effectively handles cases where the data consists of single-sentence texts.

\subsection{Theoretical Grounding}
\label{sec: Theoretical Grounding}
To provide a theoretical grounding for our obfuscation paradigm, we propose two independent constraints: the first constraint ``semantic alignment constraint'' is employed during the obfuscation generation process, while the second constraint ``Local Differential Privacy (LDP) Post-sampling constraint'' is applied post-obfuscation generations.

\paragraph{Semantic alignment constraint} Prior works on prompt obfuscation\cite{du2023dp, lyu2020towards, mai2023split, tong2023privinfer} widely adapted the concept of Differential Privacy (DP). Intuitively, this principle stipulates that: given a randomization algorithm $\mathcal{M}$, the obfuscations $\mathcal{M}(x_1)$ and $\mathcal{M}(x_2)$ of two adjacent texts $x_1$ and $x_2$ which differ by a small extent, should remain sufficiently similar, rendering them indistinguishable to adversaries \cite{chatzikokolakis2013broadening}. We adapt this principle by first defining adjacency then formulating relaxed DP under the generative paradigm.

To determine adjacency between pairs of texts (e.g., product titles, natural language clauses, etc.), we draw inspiration from \cite{przybocki2006edit} and utilize a token-level metric. 

In classic DP, two entities are considered adjacent if they differ by one instance (e.g., for two datasets, they would be adjacent if they differ by one row). Nevertheless, upon adaptation, instead of defining two texts to be adjacent if they differ by one token (which may be overly restrictive as texts must be very similar to qualify), we broaden the definition by setting a text adjacency threshold, denoted as $\rho,\, (0 \leq \rho \leq 1)$, and consider two texts as adjacent if their token-level edit-distance is $\leq \rho$ (rounded up) of the maximum token count between the two texts. Such definition allows more texts to be adjacent for privacy protection against adversaries. This relationship is defined as:

% Specifically, we calculate the token-level edit distance between the entities, requiring it to be $\leq 15\%$ (rounded up) of the maximum token count between the two entities. This relationship is formally defined as:

\begin{definition}[Text-based adjacency]
Given a pair of text $x_1, x_2$, $x_1$ and $x_2$ are adjacent if 
$$
\small
\text{ED}(x_1, x_2) \leq \\\lceil \rho \times \max(\text{TC}(x_1), \text{TC}(x_2)) \rceil 
$$
where ED denotes the token-level edit-distance, and TC represents the token count.
\end{definition}

% For final revision:
In this work, $\rho$ is set to 0.15 heuristically. In Section \ref{sec:Privacy-Utility Trade-off}, we perform a privacy-utility trade-off with different values of $\rho$. Based on this definition of adjacency, we adapt the concept of DP and propose a semantic alignment constraint for adjacent pairs within our generative obfuscation paradigm. Here, we utilize $\textsc{BertScore}$ \cite{zhang2019bertscore} to measure the semantic distance between $x_1, x_2$ and and their obfuscated counterparts $\mathcal{M}(x_1), \mathcal{M}(x_2)$.

\begin{definition}[Semantic Alignment]
Given a pair of adjacent text $x_1, x_2$, a randomization algorithm $\mathcal{M}$ is $\epsilon$-LDP under the generative paradigm if
$$\frac{\textsc{BertScore}{(x_1, x_2)}}
{\textsc{BertScore}{(\mathcal{M}(x_1), \mathcal{M}(x_2))}} \leq \epsilon 
$$
with $\epsilon$ being the privacy parameter such that $\epsilon \geq 1$.
\end{definition}

When $\epsilon = 1$, this constraint ensures that the semantic similarity between the obfuscated representations for any adjacent pair $x_1, x_2$ is at least as high as the similarity between their unobfuscated counterparts, maintaining consistency in meaning.

\paragraph{LDP Post-sampling constraint} In addition to the semantic alignment constraint, we introduce a post-sampling constraint based on a relaxed version of the privacy guarantee provided by $\epsilon$-LDP. The formal definition of $\epsilon$-LDP is as follows:

\begin{definition}[$\epsilon$-LDP]
Given two adjacent texts $x_1, x_2$ and a randomization algorithm $\mathcal{M}$, $\mathcal{M}$ satisfies $\epsilon$-LDP if for all possible outputs $y$:
$$
\quad \frac{\Pr[\mathcal{M}(x_1)=y]}{\Pr[\mathcal{M}(x_2)=y]} \leq e^\epsilon
$$
where $\epsilon$ is the privacy parameter.
\end{definition}

This condition ensures that for two similar inputs, the difference in the probability distributions of their obfuscated outputs remains bounded, thus preventing adversaries to infer the original input.

Our adaptation is designed to accommodate scenarios where the adjacent texts form a complete graph (\emph{i.e.}, each text is adjacent to every other text in the graph). In this case, we use an obfuscation LLM, LLM$_\mathcal{O}$, as $\mathcal{M}$ by generating an obfuscated representation for each of these texts, forming a set of obfuscated outputs. Instead of directly using the initial obfuscations, we apply a post-sampling technique: for each text, we sample from this set of obfuscated representations, either uniformly or based on a weighted distribution, to determine the final obfuscated output used during inference.

To illustrate, consider three adjacent texts $x_1, x_2, x_3$ in a recommendation dataset that form a complete graph. We first generate their obfuscated representations $x_1', x_2', x_3'$ using LLM$_\mathcal{O}$ and place them into a common set $S = \{x_1', x_2', x_3'\}$. If $\epsilon = 0$, each obfuscated representation is chosen with equal probability, ensuring highest privacy: for $x_1, x_2, x_3$, the probability distribution is thus 
% $\Pr[\mathcal{M}(x_i) = x_j'] = \frac{1}{3} \forall i \neq j$.  
$\Pr[\mathcal{M}(x_i) = x'_j] = \frac{1}{3}, \; \forall \, i, j \in \{1, 2, 3\}$. This uniform sampling ensures that all outputs are selected with equal likelihood, thus providing the strongest privacy guarantee based on LDP.

\begin{table*}[tt]
  \centering
  \scriptsize % This sets a smaller font size
  \begin{tabular}{l|c|c|c|c|c|c|c|c|c}
    \toprule
     & Optimization & AB (hit@10) & AT (hit@10) & MR (acc) & ES (b-acc) & CI (b-acc) & HD (b-acc) & CR (acc) & HG (cos) \\
    \hline
    GPT-4 & N/A & 0.292 & 0.415 & 0.957 & 0.949 & 0.635 & 0.648 & 0.671 & 0.755 \\
    \hline
    \hline
    SnD + GPT-4 & N/A & 0.242 & 0.352 & 0.879 & 0.881 & 0.627 & 0.635 & 0.611 & 0.684 \\
    InferDPT + GPT-4 & N/A & 0.231 & 0.339 & 0.887 & 0.868 & 0.631 & 0.641 & 0.604 & 0.692 \\
    TEP + GPT-4 & N/A & 0.219 & 0.327 & 0.853 & 0.854 & 0.619 & 0.627 & 0.589 & 0.677 \\
    \hline
    \hline
    GPT-4 + GPT-4 & Manual & \textbf{0.277} & \textbf{0.397} & 0.885 & \textbf{0.897} & \textbf{0.706} & \textbf{0.675} & \textbf{0.632} & \textbf{0.719} \\
    GPT-4 + GPT-4 & APE & \textbf{0.281} & \textbf{0.403} & 0.878 & 0.889 & \textbf{0.721} & \textbf{0.679} & \textbf{0.637} & \textbf{0.713} \\
    GPT-4 + GPT-4 & OPRO & \textbf{0.285} & \textbf{0.395} & 0.894 & 0.885 &\textbf{0.736} & \textbf{0.677} & \textbf{0.643} & \textbf{0.717} \\
    \hline
    \hline
    Gemini + GPT-4 & Manual & \textbf{0.268} & \textbf{0.379} & 0.861 & 0.871 & \textbf{0.662} & \textbf{0.683} & 0.614 & 0.698 \\
    Gemini + GPT-4 & APE & \textbf{0.271} & \textbf{0.385} & 0.857 & 0.865 & \textbf{0.671} & \textbf{0.681} & 0.619 & \textbf{0.703} \\
    Gemini + GPT-4 & OPRO & \textbf{0.273} & \textbf{0.391} & 0.874 & 0.877 & \textbf{0.674} & \textbf{0.689} & \textbf{0.622} & \textbf{0.708} \\
    \bottomrule
  \end{tabular}
  \vspace{-5pt}
  \caption{Model Performance. For $A + B$, $A$ is the obfuscation method and $B$ is the inference LLM. GPT-4 refers to inferencing with unobfuscated prompts; SnD, InferDPT, and TEP refer to the 3 obfuscation methods employed as baselines. Any output exceeding the best performance out of the 3 baseline models over $1\%$ are highlighted in bold.} 
  \label{tab:model performance}
  \vspace{-10pt}
\end{table*}

When $\epsilon > 0$, we employ the weighted sampling by setting the probability of sampling $x_i'$ for each $x_i$ proportionally higher to achieve a balance between privacy and utility. For example:
\begin{itemize}
    \small % Sets a smaller font size
    \item For $x_1, \Pr[\mathcal{M}(x_1) = x_1'] = \frac{1}{2}, \Pr[\mathcal{M}(x_1) = x_2'] = \frac{1}{4}, \Pr[\mathcal{M}(x_1) = x_3'] = \frac{1}{4}$
    \item For $x_2: \Pr[\mathcal{M}(x_2) = x_1'] = \frac{1}{4}, \Pr[\mathcal{M}(x_2) = x_2'] = \frac{1}{2}, \Pr[\mathcal{M}(x_2) = x_3'] = \frac{1}{4}$
    \item For $x_3: \Pr[\mathcal{M}(x_3) = x_1'] = \frac{1}{4}, \Pr[\mathcal{M}(x_3) = x_2'] = \frac{1}{4}, \Pr[\mathcal{M}(x_3) = x_3'] = \frac{1}{2}$
\end{itemize}

When doing so, we ensure that the probability ratios for any common obfuscated representation between adjacent entities are within the bounds specified by \( e^\epsilon \). From the above example: 

% {\footnotesize
% \[
% \frac{\Pr[M(X_1) = EX_1]}{\Pr[M(X_2) = EX_1]} = 2
% \]
% }

\vspace{-9pt}
{\small
\[
\frac{\Pr[M(x_i) = s]}{\Pr[M(x_j) = s]} \leq 2, \quad \forall \, i, j \in \{1, 2, 3\}, \, \forall \, s \in S
\]
}

This satisfies the differential privacy condition with $\epsilon = \ln(2) \approx 0.693$, making the output distributions for adjacent entities statistically similar while allowing for a controlled extent of utility in the modeling. For details on the implementation of the two constraints, please refer to Appendix \ref{sec:Constraint Implementation Specifics}.

\subsection{Obfuscation Rationale}
\label{sec: Obfuscation Rationale}
We proceed with an initial exploration into the rationale behind EmojiPrompt. To achieve this, we prompt the LLM$_{\mathcal{O}}$ to explain the reasoning for converting natural language texts into non-natural language sequences after generating obfuscated content. This explanation is conducted in two contexts: a beauty product title and a movie review, as showcased in Figure \ref{fig:encryption rationale} and the ``Obfuscation Explanation'' section of Figure \ref{fig:non-reusable-encryption}, respectively. 

As shown in Figure \ref{fig:encryption rationale}, LLM$_{\mathcal{O}}$ is capable of identifying and encoding key terms from a product title into symbolic counterparts, while altering the original syntactic structure to further obscure the content. For instance, in the obfuscated version, three droplets are introduced at the beginning, although they appear later in the original product title. This strategic reordering enhances the obfuscation, making it harder for recoverers to reconstruct the original text. Furthermore, Figure \ref{fig:non-reusable-encryption} demonstrates how LLM$_{\mathcal{O}}$ constructs non-linguistic phrases by using sequences of emojis, mathematical symbols, and logical operators to encapsulate complex expressions from a movie review. For example, a description of a ``low-budget yet commendable movie'' is represented by the emoji sequence ``Flying Money, Movie Clapper Board, OK Button.'' This exemplifies LLM$_{\mathcal{O}}$'s ability to distill intricate concepts into emblematic emoji sequences, with each emoji carrying interpretative significance. This approach enables the encoding of nuanced information while maintaining a high level of obfuscation.

% This finding also sheds light to the idea that LLMs may possess internal representations of semantic meaning or structure, as proposed by recent studies \cite{wendler2024llamas}.

These transformed sequences, as shown in both figures, introduce interpretive challenges for those attempting to recover the original content. In the case of an obfuscated beauty product description, the use of symbolic indicators may hint at certain attributes (for example, a water-drop emoji suggesting moisture), but they also introduce a level of ambiguity that prevents the clear identification of the specific product. This obfuscation is particularly effective because it conveys general information without revealing specific details. Such challenge is further amplified in the context of Non-Reusable Obfuscation, where LLM$_{\mathcal{O}}$ generates abstract symbolic sequences with nuanced, context-dependent meanings that are not immediately clear. In the future, we aim to employ more algorithms \cite{jin-etal-2025-exploring, you2024gamifying} to further analyze the rationale behind LLM’s obfuscation generations.

\begin{table*}[tt]
  \centering
  \scriptsize % This sets a smaller font size
  \begin{tabular}{l|c|c|c|c|c|c|c|c}
    \toprule
     & AB (hit@10) & AT (hit@10) & MR (acc) & ES (b-acc) & CI (b-acc) & HD (b-acc) & CR (acc) & HG (cos)  \\
    \hline
    Gemini + GPT-4 & 0.268 & 0.379 & 0.861 & 0.871 & 0.662 & 0.683  & 0.614 & 0.698 \\
    \hline
    \hline
    Gemini + GPT-4 (Content-matching) & \textbf{0.281} & 0.386 & \textbf{0.875} & 0.881 & \textbf{0.678} & \textbf{0.698}  & \textbf{0.627} & 0.702 \\
    Gemini + GPT-4 (Clause-level) & 0.265 & 0.381 & \textbf{0.883} & \textbf{0.892} & 0.659 & 0.678  & \textbf{0.631} & \textbf{0.713} \\
    Gemini + GPT-4 (Context) & \textbf{0.279} & \textbf{0.391} & \textbf{0.881} & \textbf{0.889} & \textbf{0.685} & \textbf{0.702}  & \textbf{0.638} & \textbf{0.709} \\
    \bottomrule
  \end{tabular}
  \vspace{-5pt}
  \caption{Ablation Study Performance. For $A + B$, $A$ is the obfuscation LLM and $B$ is the inference LLM. All studies use Gemini + GPT-4 as baseline, with outputs exceeding the baseline by more than $1\%$ highlighted in bold.}
  \vspace{-10pt}
  \label{tab:ablation studies}
\end{table*}

\section{Experiments}
\label{sec:experiment settings}
% In this section, we present the eight benchmarks used to evaluate the effectiveness of EmojiPrompt. These benchmarks are designed to assess the performance of our obfuscation approach across various scenarios, ensuring a comprehensive evaluation of its utility and privacy-preserving capabilities. We also detail the experimental setups and baselines used for comparison.

% In this section, we evaluate the performance of \textbf{\textbf{EmojiPrompt}} on six datasets from various fields.
\subsection{Experiment Setup}
\label{sec:Experiment Setup}

\paragraph{Dataset and Metric}
We evaluate EmojiPrompt on 8 real-world datasets from various domains where LLMs have commonly been applied to \cite{Fang2024, li2023prompt, lin2024data, rouzegar2024enhancing}: Amazon Beauty (AB), Amazon Toy (AT), Movie Review (MR), Email Spam (ES), Census Income (CI), Heart Disease (HD), Comprehensive Reading (CR), and Highlight Generation (HG). We employ Hit Rate @10 (hit@10), Accuracy (acc), Balanced Accuracy (b-acc), and Cosine Similarity (cos) as evaluation metrics. Specifically, we perform Reusable Obfuscation for AB, AT, CI, and HD, and Non-Reusable Obfuscation for MR, ES, CR, and HG. For dataset links, descriptions, modeling pre-processing, and metric details, please refer to Appendix \ref{sec:Data and Metric Specifics}.

\paragraph{Modeling Setup}
\label{sec:Modeling Setup}
To conduct a comprehensive evaluation, we implement EmojiPrompt across both trusted and untrusted LLM$_{\mathcal{I}}$ settings: \textbf{Trusted}: the LLM employed for task inference is reliable; in this case, the obfuscation only serves to prevent third-party privacy probing. \textbf{Untrusted}: the LLM employed for task inference is not reliable; that is, the obfuscation serves to prevent both third-party probing as well as potential information leakage from the server of the LLM$_{\mathcal{I}}$. In the \textbf{Trusted} scenario, we employ the same LLM for both private data obfuscation and task inference. In the \textbf{Untrusted} scenario, we employ LLMs that are hosted on distinct servers for obfuscation and inference. For both cases, our atomic-level obfuscation effectively mitigates the risk of privacy leakage, as explained in Appendix \ref{sec:Atomic Encryption Against Leakage}. In this work, we employ GPT-4 Turbo \cite{bubeck2023sparks} as the LLM$_{\mathcal{I}}$. We denote all model configurations as ``$A + B$'', with $A$ being the LLM$_{\mathcal{O}}$ and $B$ being the LLM$_{\mathcal{I}}$. Thus, in the \textbf{Trusted} scenario, GPT-4 Turbo serves both as the LLM$_{\mathcal{O}}$ and the LLM$_{\mathcal{I}}$, denoted as ``GPT-4 + GPT-4''. In contrast, for the \textbf{Untrusted} scenario, we employ Gemini 1.0 Pro \cite{team2023gemini} and Llama 3.1 (8B) \cite{vavekanand2024llama} as the LLM$_{\mathcal{O}}$s, denoted as ``Gemini + GPT-4'' and ``Llama + GPT-4'', respectively.  

% In the \textbf{Trusted} scenario, we may employ the same LLM for both privacy encryption and task inference to reduce the amount of privacy leakage under jailbreaking attacks. Conversely, in the \textbf{Untrusted} setting, we must deploy LLMs on different servers for encryption and inference, making the LLM$_{\mathcal{I}}$ unable to restore original user privacy. In this work, we employ GPT-4 Turbo as LLM$_{\mathcal{I}}$ due to its widespread use, as discussed in Section 1, with plans to explore other LLMs in future work. 

% we employ a combination of light-weighted locally hosted and cloud-based LLMs for privacy encryption — specifically, the local models are Flan-T5 Large and GPT-2 Large, denoted as ``FlanT5 + GPT-4'' and ``GPT2 + GPT-4'', and the cloud-based models are PaLM2 and Gemini-Pro, denoted as ``PaLM2 + GPT-4'' and ``Gemini + GPT-4''.

All LLMs used in this study are untuned. We apply LDP Post-sampling with $\epsilon$ = 10 on all LLM$_{\mathcal{O}}$s. We set the temperature to 1.0 for all applicable LLM$_{\mathcal{O}}$s to encourage the generation of more creative content, following \cite{roemmele2018automated}. For the LLM$_{\mathcal{I}}$, we set the temperature to 0 to obtain more consistent outputs for evaluation.

% Nevertheless, for Non-Reusable Encryption, inputting an entire text block—such as a movie review—into LLM$_{\mathcal{O}}$, regardless of whether the setting is \textbf{Trusted} and \textbf{Untrusted}, risks exposing the review to the encryption model, potentially leading to privacy breaches and server leakage. To mitigate such risks, we first segment each user review into individual sentences. Subsequently, all sentences are added to a collective corpus of review sentences. This corpus is then randomly shuffled to ensure the order of sentences does not reveal the original review context. Following this, the LLM$_{\mathcal{O}}$ is prompted to encrypt each sentence separately, converting each review sentence into its respective encrypted format. We then use the encrypted corpus of sentences to reconstruct the encrypted review for each user. In contrast, Reusable Encryption inherently avoids the depicted issue, as the entities—such as products or feature values—are not considered part of the user privacy.

\paragraph{Modeling Baselines}
\label{sec: model baselines}
We propose two types of baselines: (1) against the unobfuscated prompts, denoted as ``GPT-4'', providing a measure of how well the obfuscation retains task performance; and (2) against three prompt obfuscation models: Split-N-Denoise (SnD) \cite{mai2023split}, InferDPT \cite{tong2023privinfer}, and TokEmbPriv (TEP) \cite{qu2021natural}. For details on the baseline models, please refer to Appendix \ref{sec:Baseline Specifics}. To perform evaluation, we first obfuscate user private data within the prompts using each model, then submit the obfuscated prompts to the LLM$_{\mathcal{I}}$ (\emph{i.e.}, GPT-4 Turbo), with the configurations denoted as ``SnD + GPT-4'', ``InferDPT + GPT-4'', and ``TEP + GPT-4''.

\paragraph{Prompt Optimization}
We employ two prompt optimization algorithms, APE \cite{zhou2022large} and OPRO \cite{yang2023large}, to explore whether performance-optimized obfuscation prompts can be automatically generated, thus reducing manual effort in prompt composition. To ensure a fair comparison with the baselines, we focus solely on optimizing the obfuscation prompt, with the inference prompts fixed across all model variants. For optimization details, please refer to Appendix \ref{sec:Prompt Optimization Specifics}.

\subsection{Result and Analysis}
\label{sec: result and analysis}
As demonstrated by Table \ref{tab:model performance}, both ``GPT-4 + GPT-4'' and ``Gemini + GPT-4'' with reusable obfuscated text exhibit performance comparable and even surpassing non-obfuscated text, such as Amazon Beauty and Census Income, while showing largest relative decrease in performance on datasets with non-reusable obfuscated text (\emph{i.e.} Movie Review and Email Spam). Notably, Llama 3.1 (8B) achieves performances mostly on par with Gemini 1.0 Pro, as shown by Table \ref{tab:Llama + GPT performance} in Appendix. 

These findings shed light in the viability of using untuned LLMs as obfuscators : (1) obfuscating user private data from natural to non-natural language retains sufficient informativeness for task inference by the same LLM, and (2) the approach is extensible when using different LLMs for obfuscation and inference, as seen with ``Gemini + GPT-4'' and ``Llama + GPT-4'', though with a slight drop in performance (Appendix \ref{sec:Additional Performance Results} provides additional performance results to further validate the two observations above). Table \ref{tab:model performance} also highlights the feasibility of automatic obfuscation prompt optimization, as prompts generated by both APE and OPRO yield performance comparable to, or even surpassing, manually-tuned prompts.

\begin{table}
  \centering
  \scriptsize
  \begin{tabular}{l|c|c|c|c}
    \toprule
     & AB (hit@10) & MR (acc) & CI (b-acc) & CR (acc) \\
    \hline
    $\epsilon = 1$ & 0.251  & 0.825  & 0.639 & 0.587  \\
    $\epsilon = 3$ & 0.263  & 0.837 & 0.651 & 0.599 \\
    $\epsilon = 5$ &0.267  & 0.856 & 0.657 & 0.611 \\
    \bottomrule
  \end{tabular}
  \vspace{-5pt}
  \caption{Privacy-Utility Trade-off: Semantic Alignment}
  \label{tab:Privacy-Utility trade-off on Semantic Constraint}
  \vspace{-10pt}
\end{table}

For performance comparison with baseline models, in the \textbf{Trusted} scenario, EmojiPrompt achieves performance comparable to InferDPT on the MR dataset, while outperforming all selected baselines across other datasets. In the \textbf{Untrusted} scenario, EmojiPrompt slightly underperforms InferDPT on the MR dataset, performs comparably to SnD on both the MR and ES datasets, while outperforming all selected baselines on the remaining datasets.

% These results underscore the feasibility of employing untuned LLMs as encryptors, since: (1) encrypting user private data in the input prompt from natural language to their non-natural language equivalents can retain informativeness for task inference by the same LLM, and (2) this concept extends to scenarios where the encryption and inference LLMs differ, as in ``Gemini + GPT-4'', albeit with a decrease in performance across all evaluated datasets compared to employing the same LLM for encryption and inference.

% This finding is intriguing, as it demonstrates that our encryption paradigm is nearly fully automated, requiring minimal human involvement beyond selecting (or self-defining) the performance evaluation metrics for the prompt search. 

% Lastly, ``FlanT5 + GPT-4'', ``GPT2 + GPT-4'', and ``PaLM2 + GPT-4'' all failed at generating non-natural language encryption for corresponding private data. Please refer to Appendix \ref{sec:LLMs Incapable of Encryption} for details.

% \textit{The reason why non-reusable encrypted text have the largest relative decrease in performance may stem from the fact that we apply Non-Reusable Encryption to each sentence in the user review, transforming the entire review into a sequence of non-natural language tokens. This approach could potentially disrupt the semantic structure of the original review. In contrast, the other two tasks employ Reusable Encryption to convert each item or feature value into its corresponding non-natural language representation, with the overall semantic structure preserved.}

\subsection{Further Enhancement and Ablations}
We conduct two enhancements and one ablation study, using ``Gemini + GPT-4'' with manually-tuned obfuscation prompts as the baseline:

\textbf{Content-matching Obfuscation:} Instead of asking LLM$_{\mathcal{O}}$ to only generate obfuscated prompts $x'$, we also ask LLM$_{\mathcal{O}}$ to explain how each token in $x'$ corresponds to the original text $x$, minimizing hallucination. We denote this experiment as ``Gemini + GPT-4 (Content-matching)'' in Table \ref{tab:ablation studies}. 

\textbf{Clause-level Obfuscation:} As discussed in Section \ref{sec: Obfuscation Rationale}, in addition to token-level obfuscation, LLM$_{\mathcal{O}}$ also displays the ability to transform natural language clauses into non-linguistic sequences at the clause-level, as shown in Figure \ref{fig:non-reusable-encryption}. To explore whether this ability enhances task performance, we use all \textless \textit{natural language clause, non-linguistic sequence}\textgreater\ pairs from Figure \ref{fig:non-reusable-encryption} as in-context examples to guide generation. We denote this experiment as ``Gemini + GPT-4 (Clause-level)'' in Table \ref{tab:ablation studies}.
% \textbf{Encryption with Context}: while our atomic-level encryption helps to mitigate privacy leakage, we would like to investigate whether encrypting each entity individually without the rest of user private data as context reduces task performance. To this end, we perform a study in which we provide the LLM$_{\mathcal{O}}$ with the entire piece of user private data then prompt it to generate an encrypted representation (this practice leaks user privacy to the LLM$_{\mathcal{O}}$, and is only implemented for comparative purposes). For instance, when generating encryptions for movie reviews, instead of performing atomic-level encryption, we directly feed the entire review to the LLM$_{\mathcal{O}}$ and prompt it to return the review in encrypted form. This study is denoted as ``Gemini + GPT-4 (Context)'' in Table \ref{tab:ablation studies}.

\textbf{Obfuscation with Context:} Our atomic-level obfuscation mitigates privacy leakage yet may diminish context. To examine whether obfuscating entities individually affects task performance, we conduct a study where LLM$_{\mathcal{O}}$ is given full access to private data to generate obfuscations, allowing for a direct comparison despite privacy risks. For example, instead of obfuscating movie reviews at the atomic level, we input the full review into LLM$_{\mathcal{O}}$ for obfuscation. We denote this experiment as ``Gemini + GPT-4 (Context)'' in Table \ref{tab:ablation studies}.

As shown in Table \ref{tab:ablation studies}, Content-matching Obfuscation improves performance across all datasets. Clause-level Obfuscation boosts performance on non-reusable text datasets while maintaining similar results on others. Although atomic-level obfuscation slightly reduces task performance compared to full-context obfuscation (which leaks private data), the difference is minimal.

% \begin{table}
%   \centering
%   \scriptsize
%   \begin{tabular}{l|c|c|c|c}
%     \toprule
%      & AB (hit@10) & MR (acc) & CI (b-acc) & CR (acc) \\
%     \hline
%     $\epsilon = 1$ & 0.243 & 0.831 & 0.629 & 0.576 \\
%     $\epsilon = 3$ & 0.254 & 0.837 & 0.641 & 0.589 \\
%     $\epsilon = 5$ & \textbf{0.257} & \textbf{0.845} & \textbf{0.649} & \textbf{0.603} \\
%     \bottomrule
%   \end{tabular}
%   \caption{Privacy-Utility Trade-off: LDP Post-sampling}
%   \label{tab:Privacy-Utility trade-off on LDP Post-sampling}
% \end{table}

\subsection{Privacy-Utility Trade-off}
\label{sec:Privacy-Utility Trade-off}
% We conduct a privacy-utility trade-off evaluation for both constraints proposed in Section 3.3 using``Gemini + GPT-4'' across the Amazon Beauty, Movie Review, and Census Income datasets, with varying values of the privacy parameter. The results are presented in Table \ref{tab:Privacy-Utility trade-off on Semantic Constraint} and Table \ref{tab:Privacy-Utility trade-off on LDP Post-sampling}. Both tables showcase a monotonic drop in performance as the privacy parameter ($\epsilon$) decreases, demonstrating the effectiveness of our proposed privacy constraints.

We now conduct a privacy-utility trade-off analysis for the two constraints introduced in Section \ref{sec: Theoretical Grounding}, utilizing the "Gemini + GPT-4" configuration across the AB, MR, CI, and CR datasets with varying values of the privacy parameter ($\epsilon$). The results, as presented in Tables \ref{tab:Privacy-Utility trade-off on Semantic Constraint} and \ref{tab:Privacy-Utility trade-off on LDP Post-sampling}, reveal a monotonic decline in performance as $\epsilon$ decreases, demonstrating the effectiveness of both proposed constraints. 

Moreover, we also perform a privacy-utility trade-off analysis on the text adjacency threshold, $\rho$. The results, as shown in Table \ref{tab:Privacy-Utility trade-off on Text Adjacency Threshold}, also exhibit a monotonic drop in performances as $\rho$ increases.

% \begin{table}
%   \centering
%   \scriptsize
%   \begin{tabular}{l|c|c|c|c}
%     \toprule
%      & AB (hit@10) & MR (acc) & CI (b-acc) & CR (acc) \\
%     \hline
%     $\epsilon = 1$ & 0.243 & 0.831 & 0.629 & 0.576 \\
%     $\epsilon = 3$ & 0.254 & 0.837 & 0.641 & 0.589 \\
%     $\epsilon = 5$ & \textbf{0.257} & \textbf{0.845} & \textbf{0.649} & \textbf{0.603} \\
%     \bottomrule
%   \end{tabular}
%   \caption{Privacy-Utility Trade-off: LDP Post-sampling}
%   \label{tab:Privacy-Utility trade-off on LDP Post-sampling}
% \end{table}

\begin{table}
  \centering
  \scriptsize
  \begin{tabular}{l|c|c|c|c}
    \toprule
     & AB (hit@10) & MR (acc) & CI (b-acc) & CR (acc) \\
    \hline
    $\epsilon = 1$ & 0.243 & 0.831 & 0.629 & 0.576 \\
    $\epsilon = 3$ & 0.254 & 0.837 & 0.641 & 0.589 \\
    $\epsilon = 5$ & 0.257 & 0.845 & 0.649 & 0.603 \\
    \bottomrule
  \end{tabular}
  \vspace{-5pt}
  \caption{Privacy-Utility Trade-off: LDP Post-sampling}
  \label{tab:Privacy-Utility trade-off on LDP Post-sampling}
  \vspace{-10pt}
\end{table}

\subsection{Performance on Other Languages}
While we have demonstrated the effectiveness of our paradigm across a wide range of datasets, all datasets evaluated consist of samples composed in English. In this section, we aim to investigate whether our paradigm is generalizable to natural languages other than English. To this end, we employ four datasets, including: Spam Detection (SD, in French and German), Amazon Review Sentiment (ARS, in Japanese), Article Summarization (AS, in Chinese), and Heart Attack Detection (HAD, in Spanish). We use Cosine Similarity as the metric for Article Summarization and Balanced Accuracy for all other datasets. We adhere to the privacy settings outlined in Section \ref{sec:Experiment Setup}, while employing Gemini-1.0 Pro as the Obfuscation LLM with GPT-4 Turbo as the Inference LLM. We present the unobfuscated (denoted as ``GPT-4'') and obfuscated (denoted as ``Gemini + GPT-4'') results for all datasets in Table \ref{tab:Performance on Non-English Datasets}. As shown in Table \ref{tab:Performance on Non-English Datasets}, the obfuscated results across all datasets are comparable to their unobfuscated counterparts, thus demonstrating the effectiveness of our paradigm across various natural languages in addition to English. For dataset details, please refer to Appendix \ref{sec:Data and Metric Specifics}.  

% The unobfuscated versus obfuscated results are: 0.961 vs 0.897 (Spam in French), 0.961 vs 0.885 (Spam in German), 0.979 vs 0.942 (Amazon Review), 0.681 vs 0.642 (Article), and 0.672 vs 0.653 (Heart Attack). These results showcase the effectiveness of our method across multiple languages.

\section{Inference Attacks}
% In addition to evaluating the task performance of our encryption paradigm, we propose a comprehensive threat modeling to assess its resilience against various potential attacks. We adopt the worst-case assumption by considering the cloud-based inference LLM as untrusted. For evaluation purposes, we utilize the "Gemini + GPT-4" defined in Section 4 and analyze inference attacks targeting both the encryption LLM and the inference LLM, as the two LLMs are hosted on distinct servers with visibility into different sets of information. In this context, we define the adversary as any individual with access to the queries (or prompts) submitted to the LLM, including both the hosts of the LLM platform and third-party probers. Again, we adopt the worst-case scenario by assuming the adversary has access to all queries we submit. Please refer to Appendix \ref{sec:Thread Modeling} for proposed attacks and evaluation results.

Prior works on prompt obfuscation \cite{tong2023privinfer, yue2021differential, qu2021natural} tend to adopt token-level recovering, where the recoverer is tasked to recover each token of the privatized prompt back to its original form, with recovery accuracy reported as the evaluation metric. Nevertheless, this metric could be biased, as even if the recoverer is unable to recover the exact original tokens, it may still successfully predict synonymous tokens or generate a recovered text with a high degree of semantic similarity to the original.

To address this, we employ a comprehensive set of metrics to assess obfuscation robustness, accounting for the degree of: exact lexical overlapping (aligning with prior works), synonym and paraphrase overlapping, and overall semantic similarity between the original text and the recovered text, with both LLMs and humans as recoverers. We adopt the worst-case assumption by perceiving the cloud-based inference LLM as untrusted, and evaluate all attacks on ``Gemini + GPT-4'' against all baselines introduced in Section \ref{sec: model baselines}. For detailed descriptions on evaluation methods and results, please refer to Appendix \ref{sec:Inference Attack}. Table \ref{tab:Performance on LLM-based Attack for EmojiPrompt} in Appendix demonstrates that EmojiPrompt exhibits comparable robustness in terms of lexical overlap when benchmarked against the baselines, while achieving superior performance on both synonym overlap and overall semantic similarity.

\begin{table}
  \centering
  \scriptsize
  \begin{tabular}{l|c|c|c|c}
    \toprule
     & AB (hit@10) & MR (acc) & CI (b-acc) & CR (acc) \\
    \hline
    $\rho = 0.10$ & 0.280  & 0.881  & 0.689 & 0.629 \\
    $\rho = 0.15$ & 0.273  & 0.874 & 0.674 & 0.622 \\
    $\rho = 0.20$ & 0.256  & 0.862 & 0.667 & 0.605 \\
    \bottomrule
  \end{tabular}
  \vspace{-5pt}
  \caption{Privacy-Utility Trade-off: Text Adjacency}
  \label{tab:Privacy-Utility trade-off on Text Adjacency Threshold}
  \vspace{-12pt}
\end{table}

\section{Conclusion}
This work introduces EmojiPrompt, a novel obfuscation paradigm designed to protect user privacy during interactions with cloud-based LLMs. EmojiPrompt uses LLMs to perform generative obfuscation, transforming private data from natural language into non-natural language forms, thus obfuscating it from both LLM and human recoverers. We validate EmojiPrompt's effectiveness across eight datasets, showing that performance on obfuscated prompts is largely preserved and, in some cases, even exceeds that of unobfuscated prompts. We also compare EmojiPrompt against three obfuscation baselines, showing it matches their performance on some tasks while outperforming them in others, both for task inference and recovery robustness. Finally, the atomic-level obfuscation design allows the process to be fully cloud-based, enabling deployment without the need of local LLMs.

\section{Limitations}
We notice two potential concerns associated with employing untuned LLMs for obfuscation: 

\textbf{Limited Symbolic Vocabulary}: the restricted set of symbols—such as emojis, emoticons, and operators—from the LLM's vocabulary may constrain LLM$_{\mathcal{O}}$'s ability to fully capture the nuances of the original data. This limitation could result in the oversimplification or omission of intricate details in the obfuscated output. For instance, as shown in Figure \ref{fig:encryption rationale}, the use of a leaf emoji to denote a product's natural ingredients may not fully encapsulate the specificities of the product's organic composition. A potential solution, as \citet{edemacu2024privacy} proposes, involves expanding the symbolic vocabulary by defining additional symbol-text mappings and then incorporating these mappings into the cloud-based obfuscation LLMs, either through In-context Learning or API-based fine-tuning.

\begin{table}
  \centering
  \resizebox{0.47\textwidth}{!}{%
  \begin{tabular}{l|c|c|c|c|c}
    \toprule
       & \multicolumn{2}{c|}{SD (b-acc)} & ARS (b-acc) & AS (cos) & HAD (b-acc) \\
       & French & German & & & \\
    \midrule
    GPT-4 & 0.961 & 0.961 & 0.979 & 0.681 & 0.672 \\
    Gemini + GPT-4 & 0.897 & 0.885 & 0.942 & 0.642 & 0.653 \\
    \bottomrule
  \end{tabular}
  }
  \vspace{-5pt}
  \caption{Performance on Non-English Datasets}
  \label{tab:Performance on Non-English Datasets}
  \vspace{-12pt}
\end{table}

\textbf{Inaccurate Information}: LLMs are prone to hallucination, where they generate information that appears plausible but is factually incorrect or entirely fabricated \cite{ji2023towards}. In this work, we also observe that the obfuscation LLM has the potential to generate representations that introduce elements not present in the original data. For example, as shown in Figure \ref{fig:encryption rationale}, a sun emoji is generated for a product whose title does not specify the time of day for its use. While this may be viewed as to introduce additional noise to confuse the adversaries, it may also introduce unintended information. Several approaches may help mitigate this issue, including self-reflection \cite{ji2023towards}, knowledge distillation \cite{mcdonald2024reducing}, as well as splitting memorization and reasoning as two separated procedures \cite{jin2024disentangling}.

% $CoT paper: mitigate hallucination
% Diff layers paper: 

\bibliography{custom}

\appendix

\section{Appendix}
\label{sec:appendix}

\begin{figure*}[t]
  \centering
  \includegraphics[width=0.55\linewidth]{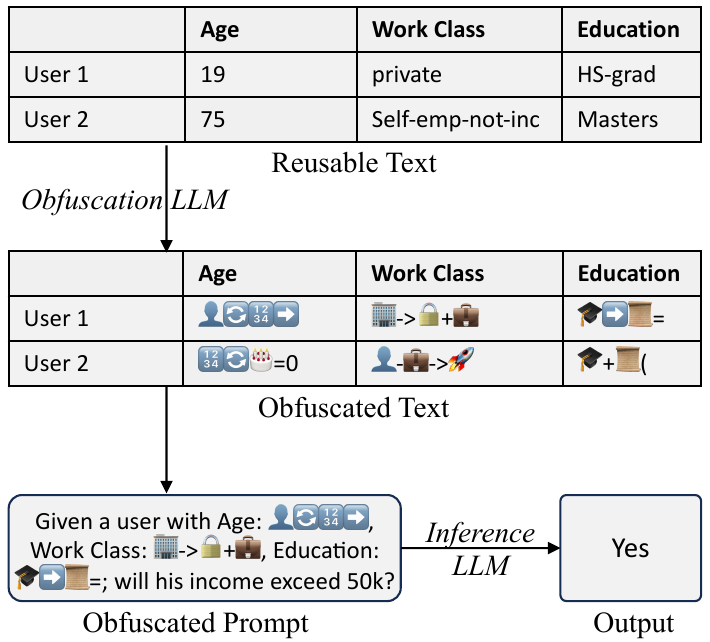}
  \captionof{figure}{Illustration of EmojiPrompt for preserving user privacy on tabular data.}
  \label{fig:model-tabular}
  % The centering effect ends here
\end{figure*}

\subsection{Obfuscation on Tabular Data} 
\label{sec:Encryption on Tabular Data}
Another example for Reusable Obfuscation is the use of LLMs for processing tabular data, a type of data commonly encountered in medical and financial decision-making. In these instances, each user is characterized by a set of predefined features (or attributes), with a specific value on each feature. For example, Figure \ref{fig:model-tabular} illustrates a simple tabular dataset with $\{$Age, Work Class, Education$\}$ as features, while each feature has two possible levels (e.g., 19 and 75 for Age). In such a scenario, the LLM$_{\mathcal{O}}$ is first used to obfuscate each level of every feature. An obfuscated user representation is then formed by aggregating all relevant obfuscated feature values, as depicted by Figure \ref{fig:model-tabular}.

Please note that for numerical features with continuous values (e.g., Age or Height), if the feature's cardinality exceeds 100, we apply quantile-based discretization to cast the cardinality to 100, and then employ the LLM$_{\mathcal{O}}$ to obfuscate each value.

\subsection{Constraint Implementation Specifics}
\label{sec:Constraint Implementation Specifics}
As inspired by prior works \cite{qu2021natural, mai2023split}, we also implement a relaxation of our semantic constraint. Specifically, we initiate by ranking all texts in an arbitrary order and proceed to sequentially obfuscate each text using an obfuscation LLM. For each text, if it has no adjacent texts, we directly assign the generated obfuscation. Otherwise, we iterate through all its adjacent texts. For each adjacent text that has been assigned an obfuscation, we compute the BertScore similarity between the newly generated obfuscation and the obfuscation of the adjacent text. To ensure that the semantic constraint is maintained, we check whether the two obfuscations retain at least $1/\epsilon$ of the original similarity between their unobfuscated counterparts. If the generated obfuscation satisfies this constraint for all its adjacent texts that have been obfuscated, we accept it and proceed to the next text. However, if the constraint is violated for any adjacent text, we waitlist the obfuscation and generate a new candidate. This process is repeated iteratively, up to a predefined number of attempts (set to 10 by default). If no valid obfuscation is found within the allowed attempts, we select the obfuscation with the highest mean semantic similarity across all its adjacent texts relative to other failed candidates. For simplicity, we compute a bidirectional BertScore to make it symmetric.

For the Post-Sampling Constraint implementation, we initiate by employing an obfuscation LLM to generate the obfuscation for each text within a task domain. Next, we construct a graph where each text is represented as a node, and two nodes are connected if the corresponding texts are considered adjacent. We then proceed by repeatedly finding the maximum clique in the graph. For each clique found, we retrieve the obfuscations of all texts within the clique and perform a post-sampling on such obfuscations with a probability distribution computed based on $\epsilon$ to obtain the final obfuscation for each node (or text) within the clique, and then remove all nodes within that clique from the graph. This process continues until the size of the maximum clique found is one.

\begin{table*}[tt]
  \centering
  \scriptsize % This sets a smaller font size
  \begin{tabular}{l|c|c|c|c|c|c|c|c|c}
    \toprule
     & Optimization & AB (hit@10) & AT (hit@10) & MR (acc) & ES (b-acc) & CI (b-acc) & HD (b-acc) & CR (acc) & HG (cos) \\
    \hline
    GPT-4 & N/A & 0.292 & 0.415 & 0.957 & 0.949 & 0.635 & 0.648 & 0.671 & 0.755 \\
    \hline
    \hline
    Llama + GPT-4 & Manual & 0.257 & 0.371 & 0.856 & 0.859 & 0.692 & 0.705 & 0.607 & 0.687 \\
    Llama + GPT-4 & APE & 0.261 & 0.376 & 0.867 & 0.851 & 0.699 & 0.701 & 0.611 & 0.691 \\
    Llama + GPT-4 & OPRO & 0.268 & 0.383 & 0.871 & 0.862 & 0.707 & 0.712 & 0.618 & 0.697 \\
    \bottomrule
  \end{tabular}
  \caption{Model Performance. For $A + B$, $A$ is the obfuscation LLM and $B$ is the inference LLM. GPT-4 refers to inferencing with unobfuscated prompts.}
  \label{tab:Llama + GPT performance}
\end{table*}

\subsection{Data and Metric Specifics}
\label{sec:Data and Metric Specifics}

\subsubsection{Dataset Description}
The Amazon \footnote{\url{https://jmcauley.ucsd.edu/data/amazon/}} datasets are collected from the Amazon.com platform with user ratings and reviews on products they have purchased on 29 categories of products. In this paper, we concentrate on evaluating performance on the Beauty and Toy categories. 

The Movie Reviews \footnote{\url{https://www.kaggle.com/datasets/lakshmi25npathi/imdb-dataset-of-50k-movie-reviews}} dataset is an aggregation of 50,000 movie reviews from the IMDB database. These reviews, composed in natural language, are evenly distributed across two sentiment classes, with 25,000 reviews categorized as positive, and the remaining 25,000 categorized as negative. 

The Email Spam \footnote{\url{https://www.kaggle.com/datasets/jackksoncsie/spam-email-dataset}} dataset consists 5,695 email messages covering a variety of topics, where each message includes the body of the email along with any associated subject lines or headers. Among these messages, 1,368 are labeled as spam.

The Census Income \footnote{\url{https://archive.ics.uci.edu/dataset/2/adult}} dataset originates from the 1994 U.S. Census database. It encompasses 14 demographic attributes per record, providing a comprehensive overview of individual census respondents. The primary target variable in this dataset is a binary indicator, signifying whether an individual's annual income exceeds \$50,000. 

The Heart Disease dataset \footnote{\url{https://www.kaggle.com/datasets/kamilpytlak/personal-key-indicators-of-heart-disease}} is originally from the CDC, which conducts annual telephone surveys to collect data on the health status of U.S. residents. It contains 18 health-related attributes per sample, with the primary target variable being whether an individual has heart disease. 

The Comprehensive Reading (Comp. Reading) dataset \footnote{\url{https://www.kaggle.com/datasets/thedevastator/introducing-quail-a-comprehensive-reading-compre?resource=download}} comprises 15,000 multiple-choice questions, balanced and annotated by question type across four distinct domains: news, user stories, fiction, and blogs. The questions are crafted to assess the reader's comprehension of the accompanying text passages. Each question has four choices.

The Highlight Generation dataset \footnote{\url{https://huggingface.co/datasets/abisee/cnn_dailymail}} contains unique news articles written by journalists at CNN and the Daily Mail, where each article is accompanied by a highlight written by the article author. The highlight has a mean token count of 56, while the article has a mean token count of 781.

For non-English datasets, the Multilingual Spam Detection data \footnote{\url{https://www.kaggle.com/datasets/rajnathpatel/multilingual-spam-data/data}} consists 5,157 unique passages in three languages, with 13\% labeled as spam. The Amazon Review Sentiment data \footnote{\url{https://github.com/tyqiangz/multilingual-sentiment-datasets/tree/main/data}} consists customer reviews in various languages. The Article Summarization data \footnote{\url{https://www.kaggle.com/datasets/noxmoon/chinese-official-daily-news-since-2016}} is from the official daily news, where each article is accompanied by a headline summarizing its content. For the Heart Attack dataset, we employ an LLM to translate all feature values from the Heart Disease dataset into Spanish.

\subsubsection{Metrics Employed}
For Amazon Beauty and Toy datasets, we use Hit@k as the evaluation metric; it computes the percentage of ranking lists that include at least one positive item in the top-K highest ranked items (we set K = 10, thus denoting it as hit@10). For the Movie Review and Comp. Reading datasets, we use accuracy as the evaluation metric, as the sentiment label and the question types are both balanced (\emph{i.e.}, with 50\% of reviews being positive, and the rest being negative). For the Email Spam, Census Income, and Heart Disease datasets, we employ Balanced Accuracy as the evaluation metric, due to the imbalanced nature of the target feature within each dataset. To clarify, Balanced Accuracy is computed as the average of the proportion of correctly predicted instances in each class, thus ensuring a fair assessment of model performance across both majority and minority classes. For the Highlight Generation dataset, we utilize Cosine Similarity as metric to quantitatively assess the semantic congruence between the target highlights (author generated) and those generated by the LLM$_{\mathcal{I}}$. 

\subsubsection{Data Processing for Modeling}
To conduct experiments on the Amazon Beauty and Toy datasets, we extract the complete purchase history for each user and retain the 15 most recently purchased items to assess the LLM's performance in sequential recommendation. Inspired by prior works \cite{geng2022recommendation,hua2023index,ji2023genrec}, we designate the most recently purchased item as the ground truth or positive sample, with the remaining 14 items as the user's interaction history. Given the finite context window size of the LLM, we adhere to the evaluation methodology proposed by \cite{geng2022recommendation, liu2023chatgpt} by randomly selecting 99 items from the entire set of beauty/toy products to serve as negative samples. These 100 sampled items collectively form a list of potential candidates for the LLM$_{\mathcal{I}}$ to rank on. Subsequently, we submit both the user's interaction history, in obfuscated form, and the sampled candidate list to the LLM$_{\mathcal{I}}$, then prompt it to generate a list of top-K recommended items for the user, based on the user's interaction history (we set K = 10).

For the Movie Reviews dataset, we furnish the LLM$_{\mathcal{I}}$ with a user's movie review in obfuscated form and instruct it to produce a binary sentiment classification (positive or negative) for the review, devoid of any accompanying explanations. A similar practice is applied to the Email Spam dataset, where we prompt the LLM$_{\mathcal{I}}$ with an obfuscated email and request the model to determine whether it is spam, also without additional explanations.

For the Census Income and Heart Disease datasets, we provide the LLM$_{\mathcal{I}}$ with obfuscated versions of each individual's attribute-value pairs. We then prompt the model to make a binary decision by responding with either ``yes'' or ``no'', without further elaboration. Specifically, the LLM$_{\mathcal{I}}$ is asked to determine whether an individual's annual income exceeds \$50,000 or whether the individual has heart disease, respectively. Given that the Census Income dataset was collected in 1994, we explicitly instruct the LLM$_{\mathcal{I}}$ to determine whether the individual's income would have exceeded \$50,000 in that year, in order to account for inflation.

For the Comp. Reading dataset, we provide the LLM$_{\mathcal{I}}$ with the privatized passage, along with the question statement as well as the four choices in natural language form. Subsequently, we prompt the LLM$_{\mathcal{I}}$ to select the most appropriate choice that answers the question based on its comprehension of the obfuscated passage. Lastly, for the Highlight Generation dataset, we provide the LLM$_{\mathcal{I}}$ with the privatized article and then prompt it to generate a highlight for the article, while ensuring that the token count of the LLM-generated highlight is less than or equal to that of the target highlight.

\subsubsection{Budget Information}
For this work, we allocate a total of \$1750 as we perform a vast amount of evaluations, covering both performance and simulated inference attacks.

\subsection{Atomic Obfuscation Against Leakage}
\label{sec:Atomic Encryption Against Leakage}
For the \textbf{Trusted} scenario, our atomic-level obfuscation helps to mitigate the risk of privacy leakage from jailbreaking attacks. Even if attackers manage to obtain one or more queries from the LLM's platform through triggering prompts (as discussed in Section \ref{sec: Introduction}), these queries remain largely uninterpretable because the user privacy is obfuscated. To accurately recover user privacy, attackers must acquire the complete set of text-obfuscation pairs for all entities in the task domain. For instance, recovering a user's purchase history would require recovering the representations of all product titles, which could number in the millions. This requirement substantially increases the difficulty of the attack compared to simply accessing a few queries.

For the \textbf{Untrusted} scenario, our atomic-level obfuscation preserves user privacy from both the obfuscation LLM and the inference LLM. For the obfuscation LLM, we prompt it to obfuscate each individual entity instead of the entire piece of private text. For instance, on product recommendation task, we prompt it to obfuscate each individual product title instead of the entire user purchase/interaction history. The rationale is that, while a user's purchase history is sensitive, the individual products are not, as they are publicly available on platforms for customers to browse. Similarly, for review (or text in natural language in general) obfuscations, the obfuscation LLM only sees the highly segmented clauses instead of the entire piece of text; such clauses can be merged in numerous ways, rendering it ambiguous on the real content of the text. Thus, even if the server host and/or third-party probers obtain such information, it would still be challenging for them to restore the user privacy. For the inference LLM, we only prompt it to perform task inferencing with user privacy in obfuscated form, so that even if the server host and/or third-party obtain such information, it would be difficult to interpret the user privacy, as they are represented in non-natural language.

\begin{table*}[tt]
  \centering
  \scriptsize % This sets a smaller font size
  \begin{tabular}{l|c|c|c|c|c|c|c|c}
    \toprule
    & AB (hit@10) & AT (hit@10) & MR (acc) & ES (b-acc) & CI (b-acc) & HD (b-acc) & CR (acc) & HG (cos) \\
    \hline
    Gemini-1.5 & 0.263 & 0.379 & 0.955 & 0.951 & 0.641 & 0.653 & 0.662 & 0.759 \\
    \hline
    \hline
    Gemini-1.5 + Gemini-1.5 & 0.258 & 0.363 & 0.887 & 0.891 & 0.719 & 0.687 & 0.645 & 0.724 \\
    GPT-3.5 + Gemini-1.5 & 0.246 & 0.349 & 0.863 & 0.872 & 0.692 & 0.681 & 0.618 & 0.703 \\
    Llama + Gemini-1.5 & 0.243 & 0.338 & 0.857 & 0.859 & 0.704 & 0.695 & 0.611 & 0.689 \\
    \bottomrule
  \end{tabular}
  \caption{Model Performance. For $A + B$, $A$ is the obfuscation LLM and $B$ is the inference LLM. Gemini-1.5 refers to inferencing with unobfuscated prompts. We employ OPRO for obfuscation prompt optimization.}
  \label{tab:Gemini-1.5 performance}
\end{table*}

\begin{table*}[tt]
  \centering
  \scriptsize % This sets a smaller font size
  \begin{tabular}{l|c|c|c|c|c|c|c|c}
    \toprule
    & AB (hit@10) & AT (hit@10) & MR (acc) & ES (b-acc) & CI (b-acc) & HD (b-acc) & CR (acc) & HG (cos) \\
    \hline
    Claude-3.5 & 0.271 & 0.386 & 0.960 & 0.949 & 0.702 & 0.711 & 0.664 & 0.753 \\
    \hline
    \hline
    Claude-3.5 + Claude-3.5 & 0.262 & 0.381 & 0.891 & 0.884 & 0.719 & 0.725 & 0.639 & 0.715 \\
    GPT-3.5 + Claude-3.5 & 0.241 & 0.369 & 0.874 & 0.869 & 0.715 & 0.731 & 0.612 & 0.699 \\
    Llama + Claude-3.5 & 0.237 & 0.362 & 0.862 & 0.853 & 0.722 & 0.727 & 0.609 & 0.684 \\
    \bottomrule
  \end{tabular}
  \caption{Model Performance. For $A + B$, $A$ is the obfuscation LLM and $B$ is the inference LLM. Claude-3.5 refers to inferencing with unobfuscated prompts. We employ OPRO for obfuscation prompt optimization.}
  \label{tab:Claude-3.5 performance}
\end{table*}

\subsection{Baseline Specifics}
\label{sec:Baseline Specifics}

For \textbf{TokEmbPriv}, we perturb the token embedding by incorporating stochastic noise prior to uploading to the server. This perturbation is implemented by introducing random noise \(Z\) drawn from a \(d\)-dimensional distribution characterized by $p(N) \propto \exp(-\eta \|N\|)$. We set the privacy parameter $\eta$ = 100 to balance utility and privacy, according to the evaluations presented in the paper. We also employ the text-to-text privatization (this post-processing procedure does not affect privacy guarantees). 

For \textbf{InferDPT}, it comprises two modules: (1) the Perturbation Module, which generates a perturbed text via \(\epsilon\)-LDP by replacing each token in the text with another from a predefined vocabulary, and (2) the Extraction Module, a locally hosted LLM (less capable than the inference LLM) that reconstructs the noisy output from the cloud-based inference LLM to better align with the original prompt.

We adhere to the methodology described in the paper by employing RANTEXT as the differential privacy mechanism, which, according to the paper, offers superior perturbation performance compared to existing state-of-the-art mechanisms. At its core, InferDPT randomly selects replacement tokens that are sufficiently close to the original token in the embedding space, with probabilities exponentially proportional to proximity. We replace $Ins_w$ with our task-specific instruction. We set \(\epsilon = 10\) to balance between utility and privacy, as demonstrated by the synonym evaluation section of the paper. As for the Extraction Module, we utilize Gemma (2B) \cite{team2024gemma} to refine open-text outputs from the inference LLM (\emph{i.e.}, outputs that are not confined to a predefined set of tokens). For Tabular datasets, we alter the sampling ranges for numerical values to enhance performances.

For \textbf{Split-N-Denoise}, 
it also adopts $d_x$-privacy to perform LDP based token-level perturbations, while introducing a novel trained local de-noising LLM to denoise outputs generated by the inference LLM with perturbed inputs. This de-noising LLM takes as input the raw user input embedding, the noise matrix, and the noisy embedding from the cloud-based inference LLM. To train the de-noising LLM in a privacy-preserving manner, a public dataset similar to the modeling dataset is needed. In this work, we utilize the validation set reserved for prompt optimization to train the local de-noising LLM. Since cloud-based inference LLMs, such as GPT-4 Turbo, require text as input, we adopt the text-to-text privatization method from TokEmbPriv, mapping perturbed token embeddings to their nearest tokens in the embedding space. We also map inference LLM outputs to their embeddings to facilitate de-noising LLM training.

% its architecture closely resembles that of InferDPT, performing LDP based token-level perturbations, while replacing the local extraction LLM with a trained local de-noising LLM. 

% This de-noising LLM takes as input the noisy embedding from the cloud-based inference LLM, the raw user input embedding, and the noise matrix. To train the de-noising LLM in a privacy-preserving manner, a public dataset similar to the modeling dataset is needed. 

% We use the same sample set reserved for prompt optimization to train the local de-noising LLM on open-text tasks. Following the paper, we set $\eta$ = 100, as the inference LLM is based on GPT. Since we do not have access to GPT-4 Turbo embeddings, we use OpenAI's ``text-embedding-3-small'' model to generate embeddings for training the local de-noiser.

Lastly, we refer to the following GitHub repositories (while modifying code if needed) for executing the baseline models: InferDPT is hosted at \url{https://github.com/mengtong0110/InferDPT}, while TokEmbPriv and Split-N-Denoise are available at \url{https://github.com/NusIoraPrivacy/eaas-privacy/tree/master}.

\subsection{Prompt Optimization Specifics}
\label{sec:Prompt Optimization Specifics}

We reserve 1,000 samples per dataset as validation set for prompt search, with the remaining data used for performance evaluation. Firstly, we manually tune the obfuscation prompt for both ``GPT-4 + GPT-4'' and ``Gemini + GPT-4''. We then employ APE and OPRO for automatic prompt optimization on the obfuscation prompt for both model variants. For both optimization algorithms, we use GPT-3.5 Turbo as the prompt generator to propose candidate obfuscation prompts based on a fixed, manually crafted meta-prompt. Also, we do not provide input-output pairs in the meta-prompt, as there is no universal ground-truth in how an entity should be obfuscated into its non-linguistic form. 

For APE, during the Monte Carlo search, 7 candidate prompts are generated per iteration. To enhance efficiency, we implement early-stopping during the evaluation of each candidate prompt and concluded the search after 6 iterations. Specifically, for every 50 samples, we compare the average performance of the current candidate prompt with that of the best-performing prompt. If the current candidate prompt underperforms the best prompt for two consecutive comparisons, we terminate its evaluation and move on to the next candidate. If the current prompt exceeds the performance of the best-performing prompt for the entire validation set, it will be updated as the new best-performing prompt. Additionally, we introduce a slight modification to the standard algorithm: instead of conducting a greedy approach that generates candidate prompts for the next iteration solely based on the best-performing prompt from the current iteration, we generate candidate prompts for the subsequent round from the top two performing prompts. 

For OPRO, we adopt the original workflow presented in the paper, where for each newly generated obfuscation prompt, we evaluate its performance score on the validation set (with early-stopping), and augment the prompt-score pair to the meta prompt. We repeat this process for up to 40 iterations, then employ the prompt with best performance. Nevertheless, we observe that initiating the search with a single obfuscation prompt may lead to repeated generation of prompts that are semantically similar to the first prompt in subsequent rounds, thus resulting in only minimal differences in performance. To resolve this, we introduce a modification by requesting the prompt generator to produce three distinct prompts and record their respective performances on the validation set. These three prompt-score pairs are then used to initiate the search process, promoting the generation of more diverse prompts in subsequent rounds.

\subsection{Additional Performance Results}
\label{sec:Additional Performance Results}

In this section, we utilize two additional cloud-based LLMs for task inferencing to evaluate whether private data obfuscated via our paradigm can be interpreted by LLMs other than GPT-4 Turbo. Specifically, we employ Gemini 1.5 Pro (denoted as Gemini-1.5) and Claude 3.5 Sonnet \cite{kurokawa2024diagnostic} (denoted as Claude-3.5). Aligning with Section \ref{sec:experiment settings}, we use the same inference LLM as the obfuscation LLM for the \textbf{Trusted} scenario, while employing GPT-3.5 Turbo (denoted as GPT-3.5) and Llama 3.1 (8B) (denoted as Llama) as obfuscation LLMs for the \textbf{Untrusted} scenario. We employ OPRO for obfuscation prompt optimization, as it has achieve best overall performance for both scenarios according to Section \ref{sec: result and analysis}. As demonstrated by Table \ref{tab:Gemini-1.5 performance} and Table \ref{tab:Claude-3.5 performance}, for both inference LLMs, obfuscating user private data from natural to non-natural language retains sufficient informativeness for task inference by the same LLM, as evidenced by the performance of ``Gemini-1.5 + Gemini-1.5'' as well as ``Claude-3.5 + Claude-3.5''. Such approach is also extensible when using different LLMs for obfuscation and inference, as demonstrated by the performance of ``GPT-3.5 + Gemini-1.5'', ``Llama + Gemini-1.5'', ``GPT-3.5 + Claude-3.5'', and ``Llama + Claude-3.5''.

\begin{table*}
    \centering
    \scriptsize
    \begin{tabular}{l|l|c|c|c|c}
        \toprule
        \textbf{} & \textbf{} & Amazon Beauty & Movie Review & Census Income & Comp. Reading  \\
        \midrule
        \multirow{4}{*}{\textbf{CosSim}} & Gemini + GPT-4 & \textbf{0.531} & \textbf{0.607} &\textbf{0.467} & \textbf{0.641} \\
        & SnD + GPT-4 & 0.609 & 0.641 & 0.518 & 0.673 \\
        & InferDPT + GPT-4 & 0.617 & 0.635 & 0.525 & 0.682 \\
        & TEP + GPT-4 & 0.622 & 0.657 & 0.534 & 0.694 \\
        \midrule
        \multirow{4}{*}{\textbf{Rouge-1}} & Gemini + GPT-4 & 0.211 & 0.183 & 0.205 & 0.228 \\
        & SnD + GPT-4 & 0.207 & 0.187 & 0.199 & 0.221 \\
        & InferDPT + GPT-4 & 0.205 & 0.193 & 0.187 & 0.219 \\
        & TEP + GPT-4 & 0.214 & 0.179 & 0.201 & 0.223 \\
        \midrule
        \multirow{4}{*}{\textbf{Rouge-2}} & Gemini + GPT-4 & 0.061 & 0.035 & 0.039 & 0.047 \\
        & SnD + GPT-4 & 0.058 & 0.037 & 0.043 & 0.051  \\
        & InferDPT + GPT-4 & 0.054 & 0.031 & 0.045 & 0.039 \\
        & TEP + GPT-4 & 0.063 & 0.042 & 0.036 & 0.045 \\
        \midrule
        \multirow{4}{*}{\textbf{Rouge-L}} & Gemini + GPT-4 & 0.207 & 0.171 & 0.198 & 0.215 \\
        & SnD + GPT-4 & 0.201 & 0.165 & 0.187 & 0.209 \\
        & InferDPT + GPT-4 & 0.198 & 0.179 & 0.191 & 0.212 \\
        & TEP + GPT-4 & 0.203 & 0.163 & 0.195 & 0.218 \\
        \midrule
        \multirow{4}{*}{\textbf{METEOR}} & Gemini + GPT-4 & \textbf{0.183} & \textbf{0.152} & \textbf{0.178} & \textbf{0.191} \\
        & SnD + GPT-4 & 0.219 & 0.193 & 0.199 & 0.238  \\
        & InferDPT + GPT-4 & 0.228 & 0.189 & 0.202 & 0.241 \\
        & TEP + GPT-4 & 0.231 & 0.196 & 0.207 & 0.247 \\
        \bottomrule
    \end{tabular}
    \caption{Recovering Robustness on LLM-based Attacks across 5 Metrics. For $A + B$, $A$ is the obfuscation method, and $B$ is the inference LLM. SnD, InferDPT, and TEP are the obfuscation methods used as baselines. Any output more than $1\%$ lower than the best performance among the baseline models (lower is better) is highlighted in bold.}
    \label{tab:Performance on LLM-based Attack for EmojiPrompt}
\end{table*}

\subsection{Inference Attack Details}
\label{sec:Inference Attack}

\subsubsection{LLM-based Attack}

In this section, we adopt from prior works \cite{tong2023privinfer, mai2023split} by proposing a simulated attack that leverages the LLM$_{\mathcal{I}}$ to recover obfuscated texts. We assume that the adversary supplies the LLM$_{\mathcal{I}}$ with details regarding the obfuscation methodology. Specifically, for EmojiPrompt, the LLM$_{\mathcal{I}}$ is informed that the obfuscation involves transforming natural language into a non-natural form using LLMs. For all baselines, the LLM$_{\mathcal{I}}$ is made aware that token-level replacements are performed, with candidate tokens chosen based on their proximity within the embedding space. The LLM$_{\mathcal{I}}$ is subsequently tasked with reversing this transformation to recover the original text from its obfuscated representation. Furthermore, we assume that the adversary provides the LLM$_{\mathcal{I}}$ with task-specific context, such as indicating that the obfuscated text corresponds to a movie review.

While prior works employ the token-level recovery rate to assess obfuscation robustness, we believe this approach may be an over-simplification. For instance, even if a recoverer fails to accurately recover the exact tokens in an obfuscated product review, it may still produce synonymous tokens or generate a recovered text that retains high semantic similarity to the original content, thus leaking the underlying meaning of the review. 

To provide a more comprehensive evaluation against this attack, we employ both semantic and token-level metrics to assess the degree of semantic similarity and lexical overlapping between the LLM-recovered text and the original text. For semantic similarity, we compute the Cosine Similarity score between the original and recovered texts, employing the ``text-embedding-3-small'' model from OpenAI for embedding vector generations, with the embedding dimension set to 200 to mitigate the curse of dimensionality following a prior work \cite{lin2024rella}. A higher similarity score would suggest lower obfuscation performance, indicating that the LLM$_{\mathcal{I}}$ can decode an obfuscated text to be more semantically similar to the original. Additionally, for token-level metrics, we compute both ROUGE (1, 2, and L) and METEOR scores between the original and recovered texts, taking into account identical tokens, synonyms, and paraphrases in the overlap. Similar to Cosine Similarity, a higher lexical overlapping would indicate that the obfuscation mechanism is less effective.

% \begin{table*}[htb]
%   \centering
%   \scriptsize
%   \begin{tabular}{l|c|c|c|c}
%     \toprule
%      & Amazon Beauty & Movie Review & Census Income & Comp. Reading\\
%     \hline
%     Cosine Similarity &0.307  &0.411  &0.295 & 0.392 \\
%     ROUGE-1 & 0.173 & 0.168  & 0.161 & 0.155 \\
%     ROUGE-2 & 0.049 & 0.019  & 0.021 & 0.017 \\
%     ROUGE-L & 0.167 & 0.152  & 0.147 & 0.149 \\
%     METEOR & 0.159 & 0.143 & 0.139 & 0.141 \\
%     \bottomrule
%   \end{tabular}
%   \caption{recovering Robustness on LLM-based Attack benchmarked against Random Entities. Random Entities involves replacing private entities in each dataset with randomly selected entities from the same dataset.}
%   \label{tab:LLM-based attack Random Entities}
% \end{table*}

We now proceed to specify how the evaluation is conducted. For reusable text, we employ the Amazon Beauty and Census Income datasets, while for non-reusable text, we utilize the Movie Review and Comp. Reading datasets. On the Amazon Beauty dataset, we present the title of each beauty product in its obfuscated form to the LLM$_{\mathcal{I}}$, prompting the model to infer the original title of the beauty product based on its obfuscated representation. We then compute the Cosine Similarity as well as ROUGE and METEOR scores between the original product title and the inferred product title. Similarly, for the Census Income dataset, we instruct the LLM$_{\mathcal{I}}$ to recover each obfuscated feature value, providing the feature name as context, and then compute the Cosine Similarity as well as ROUGE and METEOR scores between the original and recovered feature values. As for movie reviews and articles (from Comp. Reading), we direct the LLM$_{\mathcal{I}}$ to recover each obfuscated review (or article) back to its natural language form, and then compute the Cosine Similarity as well as ROUGE and METEOR scores between the original and the recovered review. For each dataset, we compute the mean scores among all entities on each metric. All scores for the EmojiPrompt and the baselines are shown in Table \ref{tab:Performance on LLM-based Attack for EmojiPrompt}.

% In contrast to most prompt encryption works on black-box LLMs that perform token-level substitutions as discussed in Section 2, our approach leverages LLMs for generative encryption. 

% Such distinction makes direct comparisons potentially biased in favor of our method. Therefore, 

In addition to benchmarking our EmojiPrompt against the 3 baseline models, we introduce a hypothetical baseline, referred to as ``Random Entities'', to enable a more comprehensive evaluation of obfuscation robustness. To establish this baseline, we randomly sample a subset of $N$ entities (we set $N = 5$) from the same dataset for each entity. We then compute the mean Cosine Similarity, ROUGE (1, 2, and L), and METEOR scores between the entity and each of the random entities. For example, in the Movie Review dataset, we randomly sample five reviews for each original review, calculate the Cosine Similarity, ROUGE, and METEOR scores between the original review and each of the five selected reviews. We then compute the mean score of the five scores across all three metrics. This procedure is repeated for all reviews in the dataset to obtain the overall mean scores for the three metrics. The same methodology is applied to all entities in the Amazon Beauty, Census Income, and Comp. Reading datasets to compute their respective overall mean scores. All scores for the ``Random Entities'' baseline are shown in Table \ref{tab:LLM-based attack Random Entities}. 

Such scores serve as a strong baseline for obfuscation performance, because if the semantic similarity (and degree of lexical overlapping) of a recovered review falls below this baseline, it indicates that the recovering LLM generates a less relevant output compared to a set of randomly selected texts, demonstrating very effective obfuscation.

As Table \ref{tab:Performance on LLM-based Attack for EmojiPrompt} showcases, when compared to the baselines, EmojiPrompt demonstrates similar robustness in lexical overlap, as evidenced by the ROUGE scores, while outperforming them in both synonym overlap and overall semantic similarity, as indicated by the METEOR and Cosine Similarity scores. While the scores are higher than those for ``Random Entities'' (as shown in Table \ref{tab:LLM-based attack Random Entities}), this is acknowledged, as the goal of the obfuscation is not to render the text entirely random but to obscure it while preserving essential information.

\begin{table*}[htb]
    \centering
    \scriptsize
    \begin{tabular}{l|l|c|c|c|c}
        \toprule
        \textbf{} & \textbf{} & Amazon Beauty & Movie Review & Census Income & Comp. Reading  \\
        \midrule
        \multirow{2}{*}{\textbf{CosSim}} & Gemini + GPT-4 & 0.531 & 0.607 & 0.467 & 0.641 \\
        & Random Entities & 0.307 & 0.411 & 0.295 & 0.392 \\
        \midrule
        \multirow{2}{*}{\textbf{Rouge-1}} & Gemini + GPT-4 & 0.211 & 0.183 & 0.205 & 0.228 \\
        & Random Entities & 0.173 & 0.168 & 0.161 & 0.155 \\
        \midrule
        \multirow{2}{*}{\textbf{Rouge-2}} & Gemini + GPT-4 & 0.061 & 0.035 & 0.039 & 0.047 \\
        & Random Entities & 0.049 & 0.019 & 0.021 & 0.017 \\
        \midrule
        \multirow{2}{*}{\textbf{Rouge-L}} & Gemini + GPT-4 & 0.207 & 0.171 & 0.198 & 0.215 \\
        & Random Entities & 0.167 & 0.152 & 0.147 & 0.149 \\
        \midrule
        \multirow{2}{*}{\textbf{METEOR}} & Gemini + GPT-4 & 0.183 & 0.152 & 0.178 & 0.191 \\
        & Random Entities & 0.159 & 0.143 & 0.139 & 0.141 \\
        \bottomrule
    \end{tabular}
    \caption{Recovering Robustness on LLM-based Attack benchmarked against Random Entities. Random Entities involves replacing private entities in each dataset with randomly selected entities from the same dataset.}
    \label{tab:LLM-based attack Random Entities}
    \vspace{-5pt}
\end{table*}

\subsubsection{Human-based Attack}

% with labmates and friends that are independent of this work as participants
In addition to utilizing an advanced LLM for inference attacks, we propose two human-based attacks on EmojiPrompt. All participants are independent of this study and were explicitly informed that their data would be used exclusively for experimental purposes within the scope of this research.

\textbf{Item Identification}: on the Amazon Beauty dataset, we randomly sample 300 item-obfuscation pairs, for each item in the pair, we pass a 500-item list (randomly sampled, with the item included) and the item obfuscation to all recoverers, asking them to identify the item from the list based on its obfuscation. We report the percentage of correctly identified items for each recoverer.

\textbf{Review Recovery}: on the Movie Review dataset, we randomly sample 300 review-obfuscation pairs, for each sampled review, we provide all recoverers with its obfuscation, asking them to recover the original review. We then report the mean Cosine Similarity, ROUGE (1, 2, and L), and METEOR scores between each human recoverer and their corresponding original reviews.

For the \textbf{Item Identification} test, five human recoverers completed the task, correctly identifying 31, 28, 24, 17, and 21 items, respectively. This results in a mean identification rate of $8.07\%$. It is important to note that while the item identification test employs a list of 500 items (including the target item) for evaluators to identify the correct item based on its obfuscation, this is already a simplified evaluation. In the domain of beauty products, there are tens of thousands of items available on major online platforms, assuming web scraping is performed, which is substantially larger than the 499 negative samples from our test. Despite this simplification, our obfuscation mechanism still demonstrates solid performance.

For the \textbf{Review Recovery} test, three human recoverers completed the task. Again, we employ the ``text-embedding-3-small'' model for embedding generation, with embedding vector dimension set to 200. The resulting mean similarity scores for the recoverers were: 0.556, 0.493, and 0.587, respectively. For ROUGE and METEOR, we report the F1 score, as it represents the harmonic mean between precision and recall. The scores for all recoverers are presented in Table \ref{tab:Human-based attack}. Overall, the performance of the human recoverers generally aligns with that of the LLM$_{\mathcal{I}}$, further underscoring the effectiveness of the obfuscation. We choose not to have the human recoverers identify the correct review from a list of candidate reviews, as we did in the Item Identification test, due to the nature of the data. In e-commerce, items are generally accessible information from popular online platforms and may be scrapped, whereas individual customer reviews are not publicly accessible unless released by the company or the individuals themselves.

\begin{table}
  \centering
  \scriptsize
  \begin{tabular}{l|c|c|c|c}
    \toprule
     & ROUGE-1 & ROUGE-2 & ROUGE-L & METEOR \\
    \hline
    Recoverer 1 & 0.189 & 0.029 & 0.169 & 0.167 \\
    Recoverer 2 & 0.178 & 0.025 & 0.153 & 0.161 \\
    Recoverer 3 & 0.191 & 0.031 & 0.178 & 0.181 \\
    \bottomrule
  \end{tabular}
  \caption{Human-based Attack for Gemini + GPT-4 on Review Recovery}
  \label{tab:Human-based attack}
  \vspace{-10pt}
\end{table}

\subsubsection{Distribution-based Attack}
% Furthermore, we introduce an additional threat model specifically for tabular datasets, where the adversary is aware of the inference task (e.g., heart disease classification) and performs web scraping to obtain public datasets closely related to the task. The adversary then examines the distribution of each value for an overlapping feature and decrypts the encrypted values based on their corresponding distributions, despite being unable to directly recover the encrypted feature values.

% For example, if the adversary obtains the distribution of 'Gender' from a public dataset similar to the target dataset, they may use this information to infer the encrypted values for 'Gender.' If there is a sufficient alignment in distribution (e.g., $70\%$ of 'Male' and $67\%$ of an encrypted value), the adversary can effectively map the encrypted values to the actual gender categories.

Furthermore, we propose a hypothetical attack aimed at tabular datasets, where the adversary targets a dataset used for a specific inference task, such as heart disease classification. Knowing the task, the adversary collects related public datasets via web scraping, which contain similar features to the target dataset. This allows them to analyze the value distributions for overlapping features.

The adversary then examines the distribution of each value in the public datasets for features that overlap with the obfuscated features in the target dataset. Despite not being able to directly recover the obfuscated feature values, the adversary uses the distribution information to make educated guesses about the obfuscated data. For example, if a public dataset shows that $70\%$ of entries for the feature ``Gender'' are ``Male'', and the obfuscated dataset has a value comprising $67\%$ of the entries for the same feature, then the adversary may infer that this obfuscated value corresponds to ``Male''.

By matching these distributions, the adversary effectively maps the obfuscated values to their actual values. This technique bypasses the obfuscation by leveraging statistical patterns rather than attempting to recover the obfuscated text directly.

To protect tabular datasets against distributional inference attacks, we propose a novel generative obfuscation mechanism with post-sampling. This approach involves generating multiple obfuscated representations for each feature value instead of mapping each value to a single obfuscation.

Consider an example with a feature that has two possible values, \(A\) and \(B\), where \(A\) accounts for 40\% of the dataset instances and \(B\) accounts for 60\%. Instead of generating a single obfuscated representation for \(A\), we prompt \text{LLM$_{\mathcal{O}}$} to produce four distinct and substantially different obfuscations: \(A_1\), \(A_2\), \(A_3\), and \(A_4\). Likewise, for \(B\), we prompt \text{LLM$_{\mathcal{O}}$} to generate three distinct and substantially different obfuscations: \(B_1\), \(B_2\), and \(B_3\). During the inference stage, one of the obfuscated representations is randomly sampled for each instance of the feature value. Consequently, for feature value \(A\), the obfuscated representations \(A_1\), \(A_2\), \(A_3\), and \(A_4\) will each appear in approximately 10\% of instances, while for feature value \(B\), the obfuscated representations \(B_1\), \(B_2\), and \(B_3\) will each appear in approximately 20\% of instances.

This method offers two key benefits. (1) by decoupling the obfuscated values from their original distribution, this mechanism prevents adversaries from accurately deducing the original feature values through distributional analysis. (2) the variability in obfuscated representations allows multiple combinations of obfuscated values to collectively approximate the original distribution proportions of 40\% for \(A\) and 60\% for \(B\). For instance, the combination of \(A_1\), \(A_2\), and \(B_1\) could collectively account for 40\% of the dataset instances. This variability makes it challenging for adversaries to ascertain which obfuscated representations correspond to specific feature values, thus obscuring discernible patterns and enhancing data security.

To assess whether our multi-obfuscation method retains task performance, we employ both Census Income and Heart Disease datasets to perform an ablation study against single obfuscated representation, with Gemini as \text{LLM$_{\mathcal{O}}$} and GPT-4 Turbo as LLM$_{\mathcal{I}}$. For each categorical feature, we randomly sample two to four unique obfuscated representations for each value of the feature. To ensure that the obfuscated representations are sufficiently different, we repeat the generation process until each obfuscated representation has a Cosine Similarity of 0.5 or less with all other representations. The balanced accuracies of multi-obfuscation versus single-obfuscation are: 0.657 vs. 0.674 (for Census Income) and 0.691 vs. 0.689 (for Heart Disease). These figures demonstrate that our method retains performance for tabular modeling.

% \subsubsection{Brute-Force Attack on Encryptor}

\end{document}